\documentclass[runningheads]{llncs}
\usepackage{cite}
\usepackage{amsmath,amssymb,amsfonts}
\usepackage{algorithmic}
\usepackage{dsfont}
\usepackage{bm}
\usepackage{textcomp}
\usepackage{xcolor}
\usepackage{soul}
\usepackage{nth}
\usepackage{graphicx}
\usepackage[misc]{ifsym}
\usepackage{mathtools}
\usepackage{graphics}
\usepackage[aboveskip=1pt]{subcaption}
\usepackage{xcolor}

\usepackage{booktabs, multirow}
\usepackage{siunitx}
\usepackage{microtype}
\usepackage{xspace}
\usepackage{latexsym}
\usepackage[pagebackref=true,breaklinks=true]{hyperref}
\hypersetup{
	colorlinks,
	linkcolor={red!75!black},
	citecolor={blue!75!black},
	urlcolor={blue!75!black},
}
\usepackage[font=small,skip=2pt]{caption}
\usepackage{siunitx}
\usepackage{microtype}
\usepackage{latexsym}
\usepackage[capitalise]{cleveref}
\crefname{section}{Sec.}{Sections}
\crefname{figure}{Fig.}{Figs.}
\crefname{table}{Tab.}{Tabs.}
\crefname{equation}{Eq.}{Eqs.}

\makeatletter
\DeclareRobustCommand\onedot{\futurelet\@let@token\@onedot}
\newcommand{\@onedot}{\ifx\@let@token.\else.\null\fi\xspace}
\newcommand{\printfnsymbol}[1]{%
  \textsuperscript{\@fnsymbol{#1}}%
}
\makeatother

\newcommand{\etal}{\emph{et al\onedot}}

\begin{document}
\title{ConFUDA: Contrastive Fewshot Unsupervised Domain Adaptation for Medical Image Segmentation}
\titlerunning{ConFUDA: Contrastive Fewshot UDA}
%


%
\author{Mingxuan gu\inst{1}\thanks{Equal contribution}, Sulaiman Vesal\inst{2}\printfnsymbol{1} (\Letter), Mareike Thies\inst{1}, Zhaoya Pan\inst{1}, Fabian Wagner\inst{1}\and Mirabela Rusu\inst{3} \and Andreas Maier\inst{1} \and Ronak Kosti\inst{1}\printfnsymbol{1}}

\authorrunning{M. Gu et al.}

\institute{
    Pattern Recognition Lab, Friedrich-Alexander-Universit\"at Erlangen-N\"urnberg (FAU), Erlangen, Germany 
    \and Department of Urology, Stanford University, Stanford CA 94305, USA 
    \and Department of Radiology, Stanford University, Stanford CA 94305, USA \\
    \email{svesal@stanford.edu}
}
\maketitle              
\begin{abstract}
Unsupervised domain adaptation (UDA) aims to transfer knowledge learned from a labeled source domain to an unlabeled target domain. Contrastive learning (CL) in the context of UDA can help to better separate classes in feature space. However, in image segmentation, the large memory footprint due to the computation of the pixel-wise contrastive loss makes it prohibitive to use. Furthermore, labeled target data is not easily available in medical imaging, and obtaining new samples is not economical. As a result, in this work, we tackle a more challenging UDA task when there are only a few (fewshot) or a single (oneshot) image available from the target domain. We apply a style transfer module to mitigate the scarcity of target samples. Then, to align the source and target features and tackle the memory issue of the traditional contrastive loss, we propose the centroid-based contrastive learning (CCL) and a centroid norm regularizer (CNR) to optimize the contrastive pairs in both direction and magnitude. In addition, we propose multi-partition centroid contrastive learning (MPCCL) to further reduce the variance in the target features. Fewshot evaluation on MS-CMRSeg dataset demonstrates that ConFUDA improves the segmentation performance by 0.34 of the Dice score on the target domain compared with the baseline, and 0.31 Dice score improvement in a more rigorous oneshot setting.

\keywords{Contrastive Learning \and Unsupervised Domain Adaptation \and  Image Segmentation \and  Deep Learning}
\end{abstract}
\section{Introduction}
\label{sec:intro}
Deep learning models are domain-sensitive and notoriously unstable when applied to a new domain with different data distributions \cite{tzeng2017adversarial}. This is more prevalent in the medical domain since the data is collected using various scanners, modalities, and acquisition parameters. Furthermore, data annotation for each modality is a time-consuming and costly process with high inter- and intra-observer variability \cite{zhuang2020cardiac}. Unsupervised domain adaptation (UDA) aims to bridge this gap by using non-linear mapping to find a common domain-invariant representation.

Recent UDA methods have used adversarial learning to align features between the source and the target domain by aligning the intermediate feature space \cite{ganin2016domain}, output space \cite{3260-08} and entropy space \cite{3260-03,Vesal2021}. Another category of works \cite{3260-05, gu2022few} utilizes real-time style transfer methods to generate artificially labeled target domain training data by transferring source domain content to the target style. 
Recently, contrastive learning has received attention in the context of UDA as it reduces domain discrepancy by forcing embedding features of the same class in different domains to be close and the ones with different classes to be apart. Singh \etal \cite{CLDA2021} compute the average of the features as the centroids and use contrastive learning to align the centroids of the same class for the source and target domains as well as to reduce the intra-domain discrepancy for classification in a semi-supervised manner. When applied to segmentation tasks with pixel-level predictions, computing the contrastive loss is more expensive than in the classification case \cite{liu2022margin}. Hu \etal \cite{Hu2021} propose block contrastive learning (BlockCL) to tackle this memory issue by dividing the features into blocks, then they apply contrastive loss only within each block. In the unsupervised scenario, positive pairs include the feature in the same coordinate in the target image and its augmented counterpart. The features of a different coordinate form the negative pairs. Since it is possible for features of the same class to be included in the negative pairs, the segmentation module can be easily confused.

We propose a contrastive fewshot UDA approach called \textbf{ConFUDA} for multi-modal cardiac image segmentation. Fewshot UDA \cite{3260-05} is a more challenging but realistic scenario with very few unlabeled target samples available for training. To overcome the computational burden of contrastive learning for UDA applied to image segmentation, we introduce a novel notion of centroid computation for contrastive pairs. 

\textbf{Our main contributions} are four-fold: (1)~We adopt a style transfer module (RAIN \cite{3260-05}) to progressively generate more difficult target-like images preserving the content of source samples for the segmentation module; (2)~We propose a centroid-based contrastive learning approach (CCL) for image segmentation to relieve the computational burden of contrastive learning and a centroid norm regularizer (CNR) to restrict centroids in magnitude as a complement of the contrastive loss; (3)~We propose a multi-partition centroid contrastive learning (MPCCL) to further reduce the variance of the features within the same class; and (4)~We validate our UDA approach on the rigorous scenario when there is only one or few target data available.

\section{Methods}
In this section, we present the novel contrastive fewshot unsupervised domain adaptation (ConFUDA).We first introduce the notations, then we describe each component of ConFUDA in detail.

\noindent\textbf{Problem statement:} 
In medical image segmentation under UDA setting, we are given a set of labeled source samples $\mathcal{D}^s=(\{x_i^s\}_{i=1}^{N^s}, \{y_i^s\}_{i=1}^{N^s})$; and a set of unlabeled target samples $\mathcal{D}^t=(\{x_i^t\}_{i=1}^{N^t})$, where $N^s$ and $N^t$ are the number of source and target samples respectively. The goal of UDA segmentation is to improve performance of the model on $\mathcal{D}^t$ by utilizing the labeled $\mathcal{D}^s$ and aligning the distribution between $\mathcal{D}^s$ and $\mathcal{D}^t$. In this work, we consider a more restrictive setting where only one or few unlabeled target samples are used during training.

\begin{figure}[bt]
 \includegraphics[width=\textwidth]{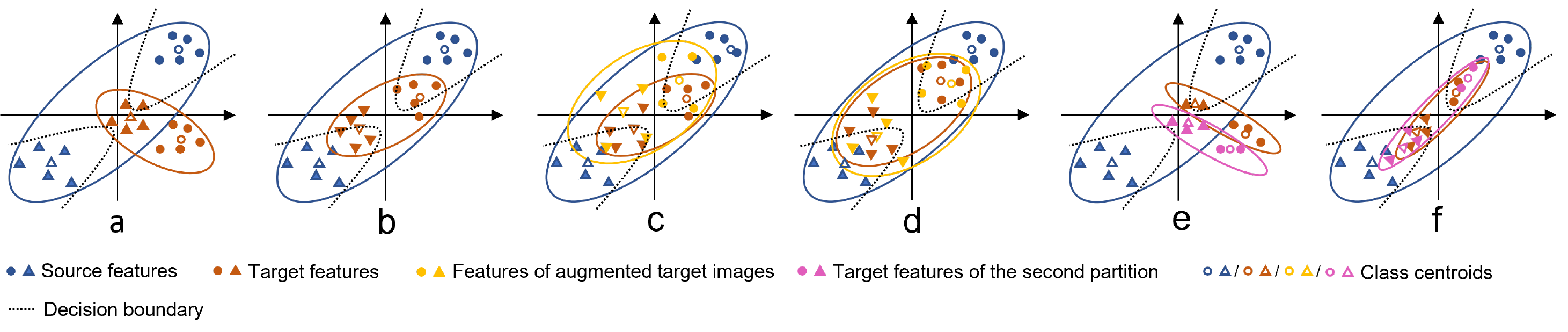}
\caption{Illustration of the proposed ConFUDA. (a-b) 
Inter-domain contrastive learning aligns the feature centroids from both domains. (c-d) 
Intra-domain contrastive learning helps to move the features away from the decision boundary. (e-f) 
MPCCL aligns the centroids between each partition and source centroids and makes the target features more compact.} 
\label{fig:dot_graph}
\end{figure}

\subsection{Random Adaptive Instance Normalization (RAIN)}\label{subsec:rain}

RAIN \cite{3260-05}, a style transfer module, uses Adaptive Instance Normalization (AdaIN) \cite{3260-07} as its basis and takes two images to generate a new image that consists of the content of one image and the style of the other. RAIN extends AdaIN and uses a variational auto-encoder (VAE) to create a style distribution of a single image. A normal distributed variable $\epsilon$ in the VAE can be used to tune the generated stylized image. As a result, by updating the $\epsilon$ in the negative direction of the gradient propagated from the main task, RAIN is able to generate more difficult stylized images progressively, thus helping the model become more robust. The loss function for RAIN is: $\mathcal{L}_{RAIN} = \mathcal{L}_{c} + \lambda_{s}\mathcal{L}_{s} + \lambda_{KL}\mathcal{L}_{KL} + \lambda_{Rec} \mathcal{L}_{Rec}$,
where $\mathcal{L}_{c}$ and $\mathcal{L}_{s}$ are content and style losses for AdaIN to generate realistic stylized images, $\mathcal{L}_{KL}$ and $\mathcal{L}_{Rec}$ are KL-divergence and reconstruction losses applied to the VAE to enforce normal distributed statistics and an identity reconstruction of the VAE. Similar to FUDA\cite{gu2022few}, we set the weights of the losses $\lambda_s=5$, $\lambda_{KL}=1$ and $\lambda_{Rec}=5$ to train the RAIN module.

\begin{figure}[bt]
\centering
 \includegraphics[width=\textwidth]{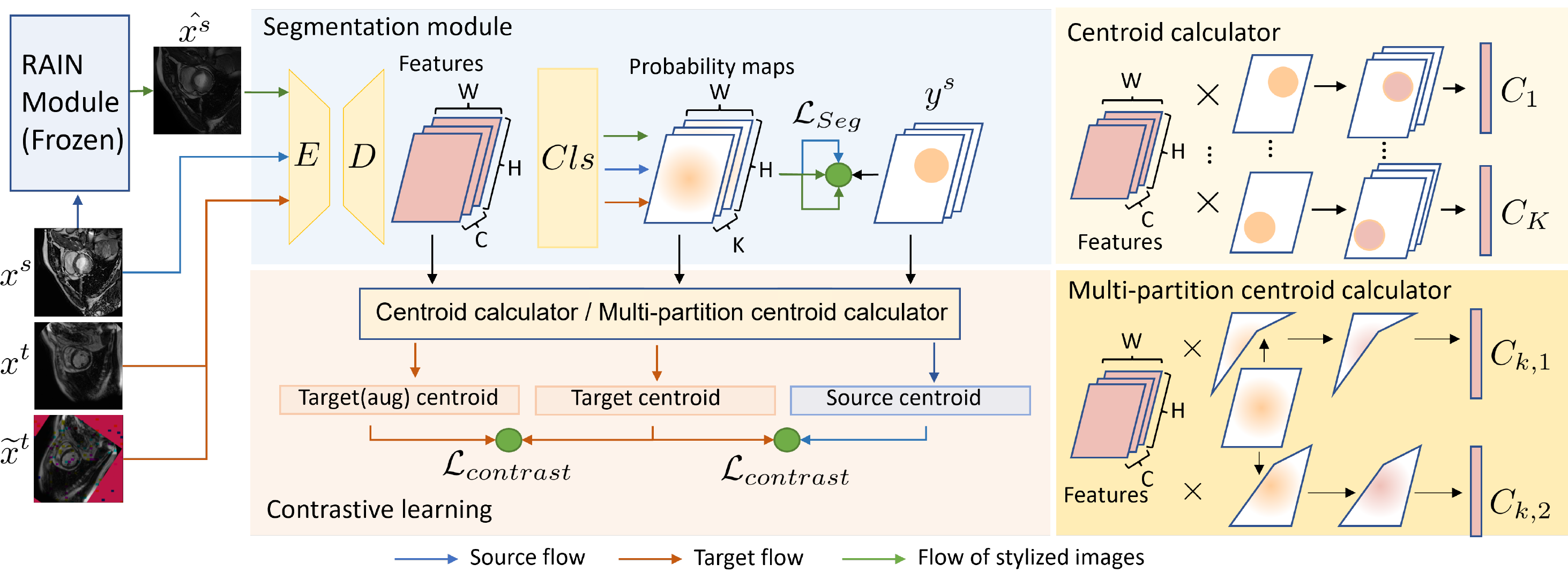}
\caption{Overview of our ConFUDA network architecture, where $E$, $D$ and $Cls$ represent encoder, decoder, and classifier of the segmentation network. The classifier $Cls$ is simply a 1$\times$1 convolutional layer generating the prediction from the decoder feature. We apply RAIN to generate target-stylized source images $\hat{x}^s$.
Then, $x^s$ and $\hat{x}^s$ are used to train the segmentation module. 
To perform feature alignment with contrastive learning, $\hat{x}^s$, $x^s$, target image $x^t$ and its augmented counterpart $\widetilde{x}^t$ are fed into the segmentation module; and then the centroid calculator or multi-partition centroid calculator is used to generating the corresponding class-wise centroids for each image. Contrastive losses are then computed with the generated centroids during training.} 
\label{fig:architecture}
\end{figure}

\subsection{Centroid-based Contrastive Learning (CCL)}\label{subsec:contrastive}
As mentioned in \cref{sec:intro}, Singh \etal \cite{CLDA2021} proposed contrastive learning for semi-supervised domain adaptation (CLDA) to tackle the domain adaptation problem for classification by computing the means of the encoder features as class-wise centroids, and applying contrastive losses between centroids of different domains. For the classification task in CLDA, each image only represents one class, while in image segmentation there are multiple class structures in each image. Averaging the features will fuse the features of different classes together. Therefore, CLDA cannot be applied directly to image segmentation. To solve this problem, taking source images as an example, we mask out the decoder features by multiplying the features with their binary labels $y^s$ (0, 1). Then, we compute the means of the features for each class in spatial dimension to generate class-wise centroids for each class, as is illustrated in \cref{fig:architecture} and can be written as:
 
\begin{equation}
    C_{k}^{s} = \frac{\sum_{b=1}^{B}\sum_{h,w=1}^{H,W}(D(E(x^s))*y_{k}^s)}{\sum_{b, h, w=1}^{B, H, W} y_{k}^s}
    \label{eq:centroid}
\end{equation}
where, $k \in \{1, 2, ..., K\}$ is a class indicator, $B$ is the batch size and $H, W$ denote the height and width resolution. Similar to CLDA, we also maintain a memory bank $(C^s = [C_1^s, C_2^s, ..., C_K^s])$. During training, we update source centroids with exponential moving average:
    $C_{k}^{s} = \rho(C_{k}^{s})_{step} + (1-\rho)(C_{k}^{s})_{step-1},$
where $\rho=0.9$.

For target images $x^t$, soft prediction $Softmax(Cls(D(E(x^t))))$ of $x^t$ are used as the pseudo label to softly mask out the decoder features of $x^t$. Then, the centroids of target images can be calculated following the \cref{eq:centroid}.

\noindent\textbf{Inter- and Intra-domain Contrastive Learning.}
To align the class-wise centroids of source and target domain, we employ the inter-domain contrastive loss as a modified NT\_Xent (normalized temperature-scaled cross-entropy loss) \cite{chen2020simple} taking centroids of the same class $C_{k}^t$, $C_{k}^s$ as the positive pairs and centroids of different classes as negative pairs:
\begin{equation}
    \mathcal{L}_{inter} = \mathcal{L}_{contrast}(C^{t}, C^{s}) = -\sum_{k=1}^{K} log\frac{h(C_{k}^{t}, C_{k}^{s})}{\sum_{d\in {s, t}}\sum_{i=1, i\neq k}^{K}h(C_{k}^{t}, C_{i}^{d})}
    \label{eq:contrast}
\end{equation}
where $h(u, v) = exp(\frac{u^Tv}{\left \| u \right \|_2\left \| v \right \|_2} / \tau)$ measures the exponential of cosine distance between two feature vectors, $\tau$ is the temperature and $K$ is the number of classes.
Inter-domain contrastive learning pulls the source and target features of the same class closer, while target features close to the decision boundary may be pushed to the wrong class after adaptation\cite{CLDA2021} (\cref{fig:dot_graph}c). To solve this problem, we adapt the idea behind the instance contrastive alignment in CLDA and propose the intra-domain contrastive learning. We sample the target images with heavy augmentations. Then, we compute their respective centroids ($C^t$, $\widetilde{C}^t$) and apply a contrastive loss $\mathcal{L}_{contrast}(C^t, \widetilde{C}^t)$. As shown in \cref{fig:dot_graph}(c-d), to align the target features with their augmented counterparts, the model has to push both features to a high-density region, which leaves the decision boundary in the low-density region, thus improving the performance of the model.

\noindent\textbf{Centroid Norm Regularizer (CNR).}
As the centroids are normalized to unit vectors in contrastive loss, it can only optimize the direction of the centroid and has no control over the magnitude of the centroids. To bring the source and target centroids ($C^s$, ${C}^t$) even closer, we propose the centroid norm regularizer (CNR) to regularize the magnitude of the target centroids taking source centroids as the labels. In practice, we compute the L2 distance loss between the norms of the centroids which is formulated as:
$    \mathcal{R}(C^{t}) = \frac{1}{K}\sum_{k=1}^{K} (\left \| C_{k}^{t} \right \|_2 - \left \| C_{k}^{s} \right \|_2)^2 $

\noindent\textbf{Multi-partition Centroid Contrastive Learning (MPCCL).}\label{subsec:mpccl}
During training, we observed vanishing gradient problem on the decoder features, as during optimization of centroids, the small gradients of the centroids cause the gradients propagated to the features to be even smaller. Therefore, when the centroids are optimized, the features will not be updated anymore. To mitigate this problem, we propose a simple and efficient method called MPCCL. As illustrated in \cref{fig:dot_graph} (e-f), we first equally and randomly split the soft prediction of target images into $P$ partitions. Then, reusing \cref{eq:centroid} we compute the centroids for each partition and apply inter- and intra-domain contrastive loss for each partition's centroid. The final contrastive loss can be written as:
\begin{equation}
    \mathcal{L}_{p\_contrast}(C^{s}, C^{t}, \widetilde{C}^{t}) = \frac{1}{P}\sum_{p=1}^{P}(\mathcal{L}_{contrast}(C_{p}^{t}, C^{s}) + \mathcal{L}_{contrast}(\widetilde{C}^t, C_{p}^{t}))
\end{equation}
To show the intuition behind MPCCL, by increasing the number of partitions from $P$ to $H \times W$, we can derive the traditional contrastive loss from MPCCL. Thus, MPCCL can be considered as a compromise between the large computational burden and the model performance.

Combining all the components (\cref{subsec:rain}--\cref{subsec:contrastive}) 
, we state the overall loss function in \cref{eq:totalloss}. Here, $\lambda_{contrast}$ and $\lambda_{CNR}$ are the weights for the corresponding losses, and $\mathcal{L}_{Seg}$ is a combination of cross-entropy and Jaccard loss.  
\begin{equation}
    \mathcal{L} = \mathcal{L}_{RAIN} + \mathcal{L}_{Seg} + \lambda_{contrast}\mathcal{L}_{p\_contrast}(C^s, C^t, \widetilde{C}^t) + \lambda_{CNR}\sum_{p=1}^{P}\mathcal{R}(C_{p}^t)
    \label{eq:totalloss}
\end{equation}

\section{Experiment Results}
\subsection{Dataset}
To evaluate the proposed ConFUDA, we used the MS-CMRSeg\cite{zhuang2020cardiac} challenge dataset. The dataset contains 45 short-axis bSSFP, T2-weighted, and LGE scans from patients diagnosed with cardiomyopathy. Three structures are included: the right ventricle (RV) cavity, the left ventricle (LV) cavity, and the myocardium (MYO) region. Only affine transformations are applied to the bSSFP (source) images. Heavy augmentations for LGE (target) images include affine transformation, elastic transformation, Gaussian noise, random drop out, and gaussian blur. The sequences are normalized with min-max normalization. All the images are center cropped to 224 $\times\ 224$ pixels on the region of interest areas.

\begin{table}[!tb]
\centering
\caption{The average Dice coefficient (Avg. Dice) and average Hausdorff distance (Avg. HD95) measures of 5-fold cross-validation for ConFUDA together with the baseline (W/o UDA), inter-observer study, and domain adaptation (AdaptSeg, AdvEnt, FUDA) and contrastive learning (BlockCL) methods. Results are shown in $mean\pm std$ by computing the average of the $mean$ and $std$ across 5 folds. For ``Oneshot'', only one slice of the target image is used for training. For ``Fewshot'', we take a sequence of target images as the target training data. All of the target training data are used for ``full''. The best results are highlighted in bold.}

\resizebox{\textwidth}{!}{%
\begin{tabular}{lc|ccc|c|ccc|c}
\toprule
\textbf{Exp.} & \textbf{Method} & \textbf{MYO} & \textbf{LV} & \textbf{RV} & \textbf{Avg. Dice} & \textbf{MYO} & \textbf{LV} & \textbf{RV} & \textbf{Avg. HD95} \\
\midrule
  N/A   & W/o UDA        & 0.24$\pm$0.28     & 0.39$\pm$0.35      & 0.51$\pm$0.33      & 0.38$\pm$0.32 & 31.7$\pm$19.1     & 24.7$\pm$22.8     & 28.0$\pm$20.2  & 28.1$\pm$20.7         \\
\midrule
N/A    &Observer        & 0.76$\pm$0.07 & 0.07$\pm$0.88 & 0.88$\pm$0.06 & \textbf{0.82$\pm$0.08}   & 0.1$\pm$0.8   & 0.8$\pm$0.1   & 0.1$\pm$12.0    & 12.0$\pm$4.4 \\
\midrule
\multirow{5}{*}{\rotatebox{90}{Oneshot}}    & AdaptSeg  & 0.35$\pm$0.27 & 0.6$\pm$0.29  & 0.52$\pm$0.28 & 0.49$\pm$0.28            & 18.9$\pm$25.9 & 13.2$\pm$16.6 & 23.4$\pm$31.5 & 18.5$\pm$24.7         \\
                                              & AdvEnt  & 0.39$\pm$0.27 & 0.61$\pm$0.30  & 0.57$\pm$0.27 & 0.52$\pm$0.28            & 15.7$\pm$21.8 & 13.8$\pm$17.1 & 18.3$\pm$24.0 & 15.9$\pm$21.0       \\
                                              & BlockCL & 0.58$\pm$0.15       & 0.81$\pm$0.11       & 0.63$\pm$0.23       & 0.67$\pm$0.16                  & 20.0$\pm$14.2       & 10.7$\pm$7.8       & 17.5$\pm$17.3       & 16.0$\pm$13.1         \\
                                              & FUDA & 0.48$\pm$0.19    & 0.80$\pm$0.12    & 0.67$\pm$0.24    & 0.65$\pm$0.18               & 13.2$\pm$20.1    & 10.0$\pm$13.5   & 14.3$\pm$23.8    & 12.5$\pm$19.4 \\
                                              & ConFUDA & \textbf{0.55$\pm$0.15} & \textbf{0.84$\pm$0.08}  & \textbf{0.69$\pm$0.24}  & \textbf{0.69$\pm$0.16}            & \textbf{7.8$\pm$5.7} & \textbf{7.6$\pm$6.5} & \textbf{11.0$\pm$18.2} & \textbf{8.8$\pm$10.1} \\
                                              \midrule
\multirow{5}{*}{\rotatebox{90}{Fewshot}}      & AdaptSeg  & 0.42$\pm$0.24 & 0.65$\pm$0.25 & 0.54$\pm$0.26 & 0.54$\pm$0.25            & 14.6$\pm$16.6 & 13.2$\pm$16.0   & 18.8$\pm$28.0   & 15.5$\pm$20.2       \\
                                              & AdvEnt  & 0.40$\pm$0.25  & 0.62$\pm$0.27 & 0.63$\pm$0.23 & 0.55$\pm$0.25            & 14.8$\pm$19.7 & 14.8$\pm$24.0 & 15.4$\pm$21.1 & 15.0$\pm$21.6       \\
                                              & BlockCL &  0.56$\pm$0.17       & 0.80$\pm$0.11       & 0.66$\pm$0.24       & 0.67$\pm$0.17   & 13.1$\pm$10.1       & 14.8$\pm$11.5       & 13.0$\pm$18.9       & 13.6$\pm$13.5       \\
                                              & FUDA & 0.50$\pm$0.17    & 0.81$\pm$0.11    & 0.70$\pm$0.23    & 0.67$\pm$0.17               & 10.7$\pm$13.4    & 8.5$\pm$9.1 & 13.3$\pm$24.2    & 10.8$\pm$15.6 \\
                                              & ConFUDA & \textbf{0.58$\pm$0.14} & \textbf{0.85$\pm$0.07} & \textbf{0.72$\pm$0.20}  & \textbf{0.72$\pm$0.13}            & \textbf{7.6$\pm$9.4} & \textbf{7.2$\pm$6.0} & \textbf{10.2$\pm$15.1} & \textbf{8.3$\pm$10.2} \\
                                              \midrule
\multirow{5}{*}{\rotatebox{90}{Full}} & AdaptSeg  & 0.43$\pm$0.22 & 0.64$\pm$0.24 & 0.66$\pm$0.18 & 0.58$\pm$0.21            & 12.8$\pm$13.3 & 15.2$\pm$14.0   & 10.9$\pm$08.8   & 12.9$\pm$12.0             \\
                                              & AdvEnt  & 0.45$\pm$0.23 & 0.66$\pm$0.25 & 0.67$\pm$0.20 & 0.59$\pm$0.23            & 11.3$\pm$11.0 & 9.2$\pm$7.3 & 9.4$\pm$9.9 & 10.0$\pm$9.4             \\
                                              & BlockCL & 0.52$\pm$0.21       & 0.83$\pm$0.08       & 0.71$\pm$0.19       & 0.68$\pm$0.16                  & 11.1$\pm$13.0       & 7.9$\pm$5.8       & 10.2$\pm$14.1       & 9.7$\pm$11.0             \\
                                              & FUDA & 0.51$\pm$0.18       & 0.82$\pm$0.11       & 0.68$\pm$0.23       & 0.67$\pm$0.18                  & 7.8$\pm$6.3       & \textbf{6.3$\pm$3.8}       & 11.9$\pm$19.6       & 8.7$\pm$9.9 \\
                                              & ConFUDA & \textbf{0.62$\pm$0.11}                         & \textbf{0.84$\pm$0.07}                         & \multicolumn{1}{c|}{\textbf{0.73$\pm$0.15}}                         & \multicolumn{1}{c|}{\textbf{0.73$\pm$0.11}}                         & \textbf{6.6$\pm$4.5}                         & 10.5$\pm$8.7                           & \multicolumn{1}{c|}{\textbf{8.0$\pm$8.2}} & \textbf{8.4$\pm$7.2} \\ \bottomrule
\end{tabular}}
\label{tab:comparision}
\end{table}
\begin{figure}[!tb]
     \centering

     \begin{subfigure}[b]{0.13\textwidth}
         \centering
         \includegraphics[width=\textwidth, height=0.07\textheight]{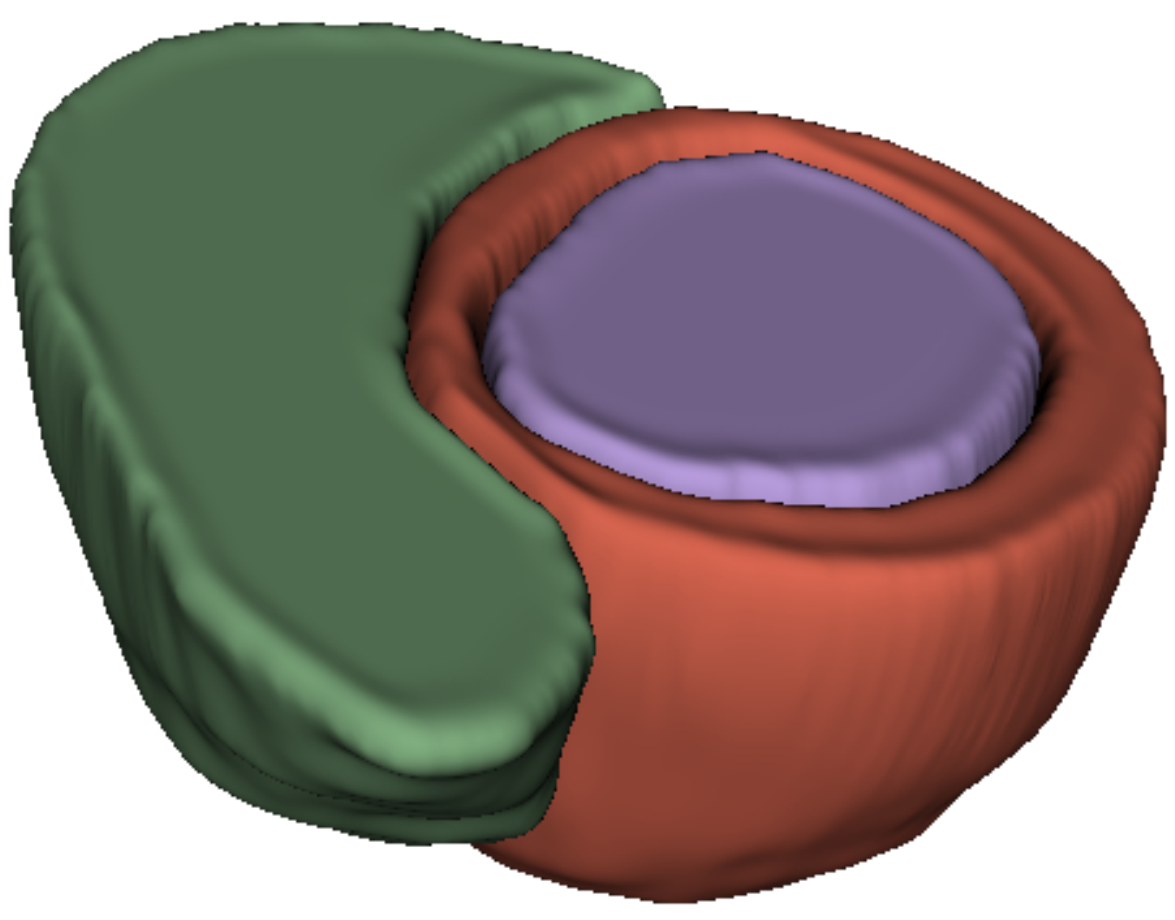}
         \label{}
     \end{subfigure}
     \begin{subfigure}[b]{0.13\textwidth}
         \centering
         \includegraphics[width=\textwidth, height=0.07\textheight]{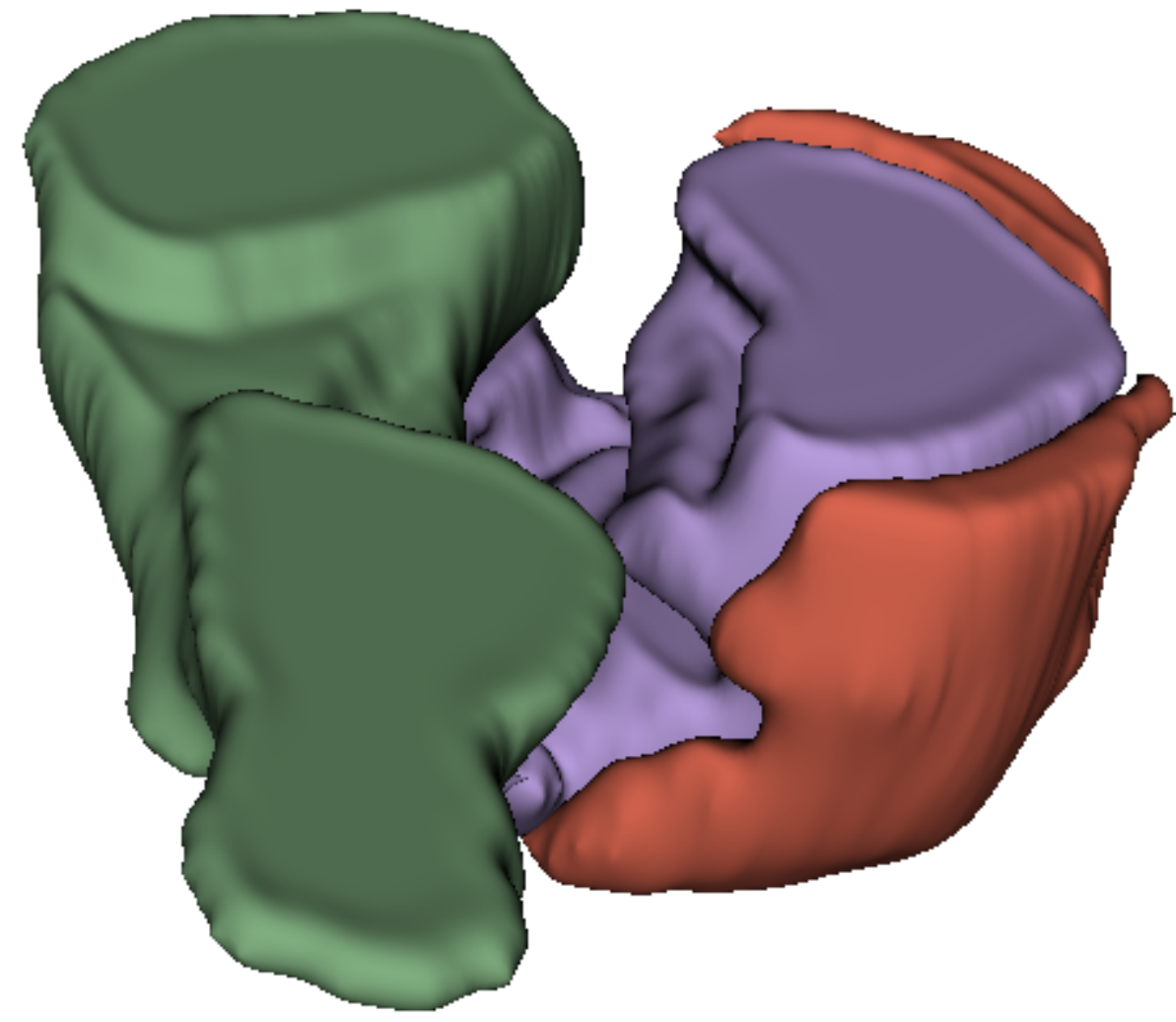}
         \label{}
     \end{subfigure}
    \begin{subfigure}[b]{0.13\textwidth}
         \centering
         \includegraphics[width=\textwidth, height=0.07\textheight]{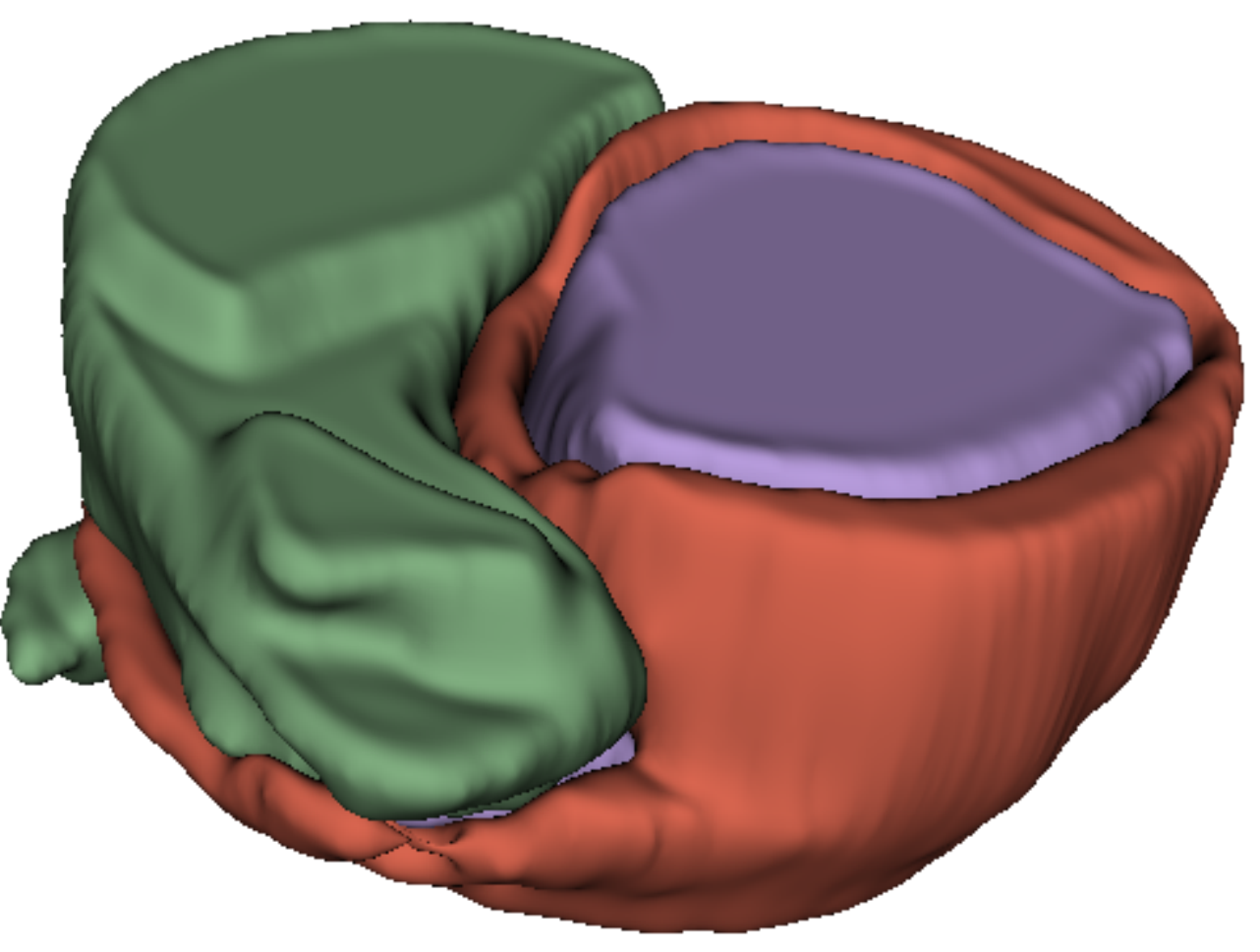}
         \label{}
     \end{subfigure}
    \begin{subfigure}[b]{0.13\textwidth}
         \centering
         \includegraphics[width=\textwidth, height=0.07\textheight]{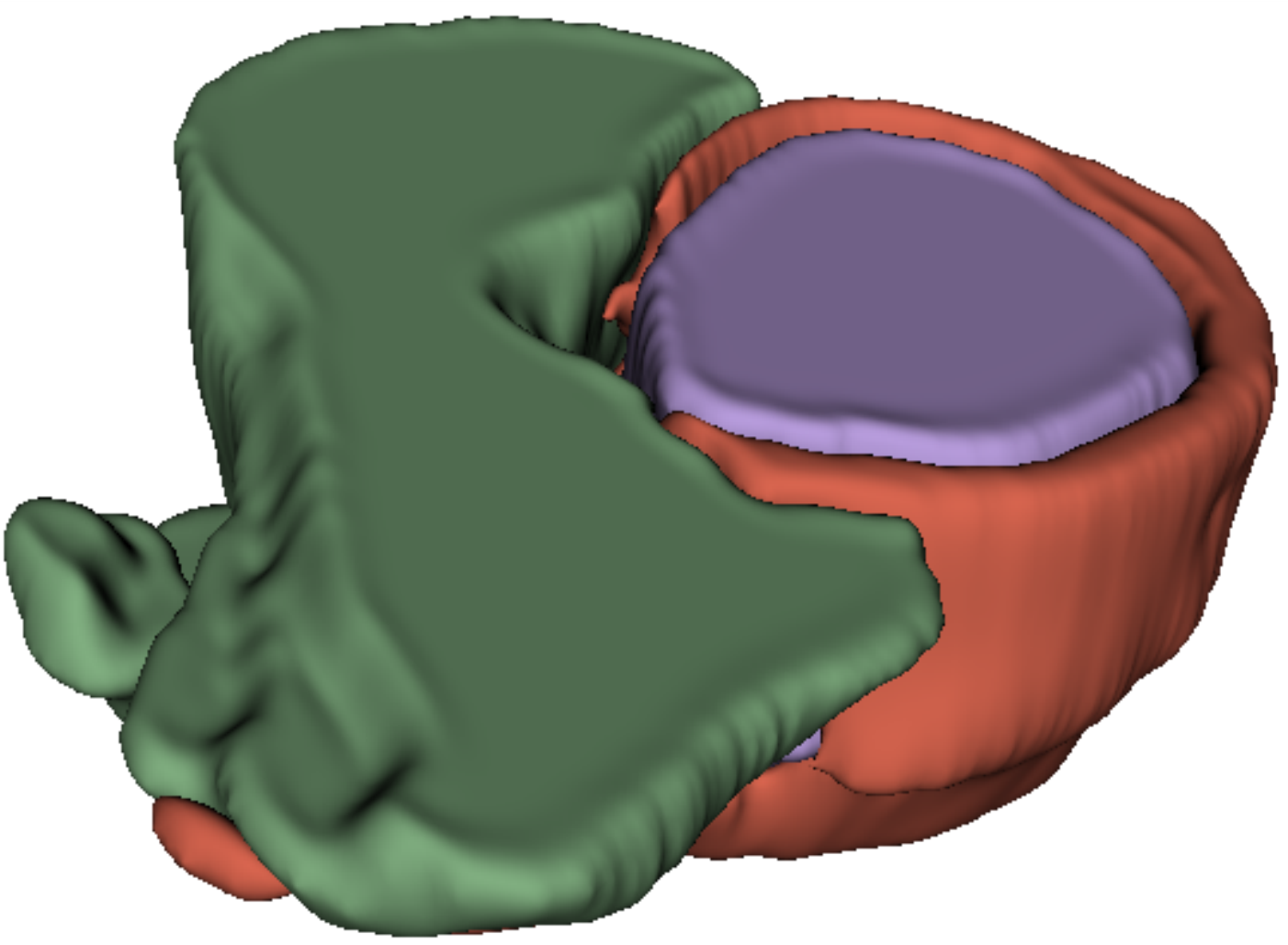}
         \label{}
     \end{subfigure}
    \begin{subfigure}[b]{0.13\textwidth}
         \centering
         \includegraphics[width=\textwidth, height=0.07\textheight]{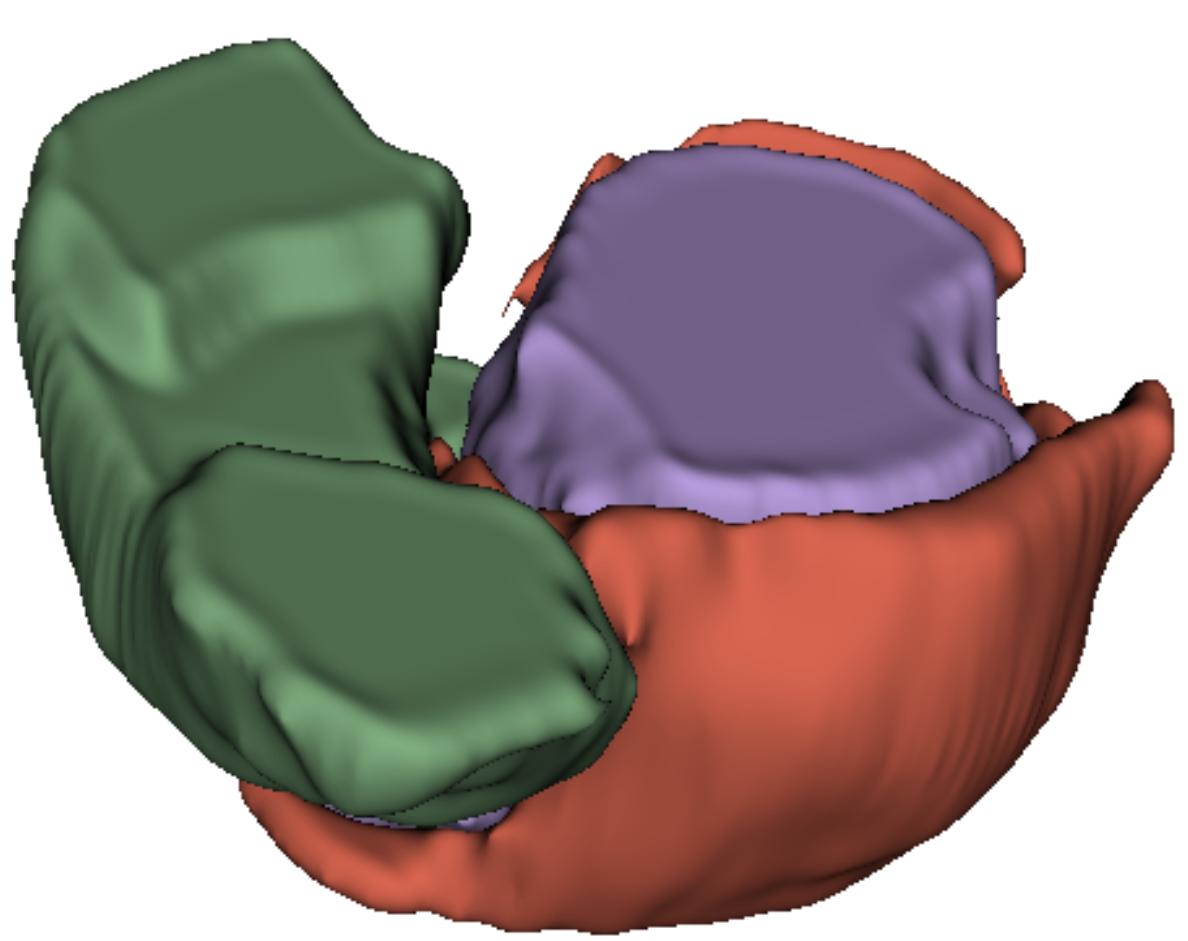}
         \label{}
     \end{subfigure}
    \begin{subfigure}[b]{0.13\textwidth}
         \centering
         \includegraphics[width=\textwidth, height=0.07\textheight]{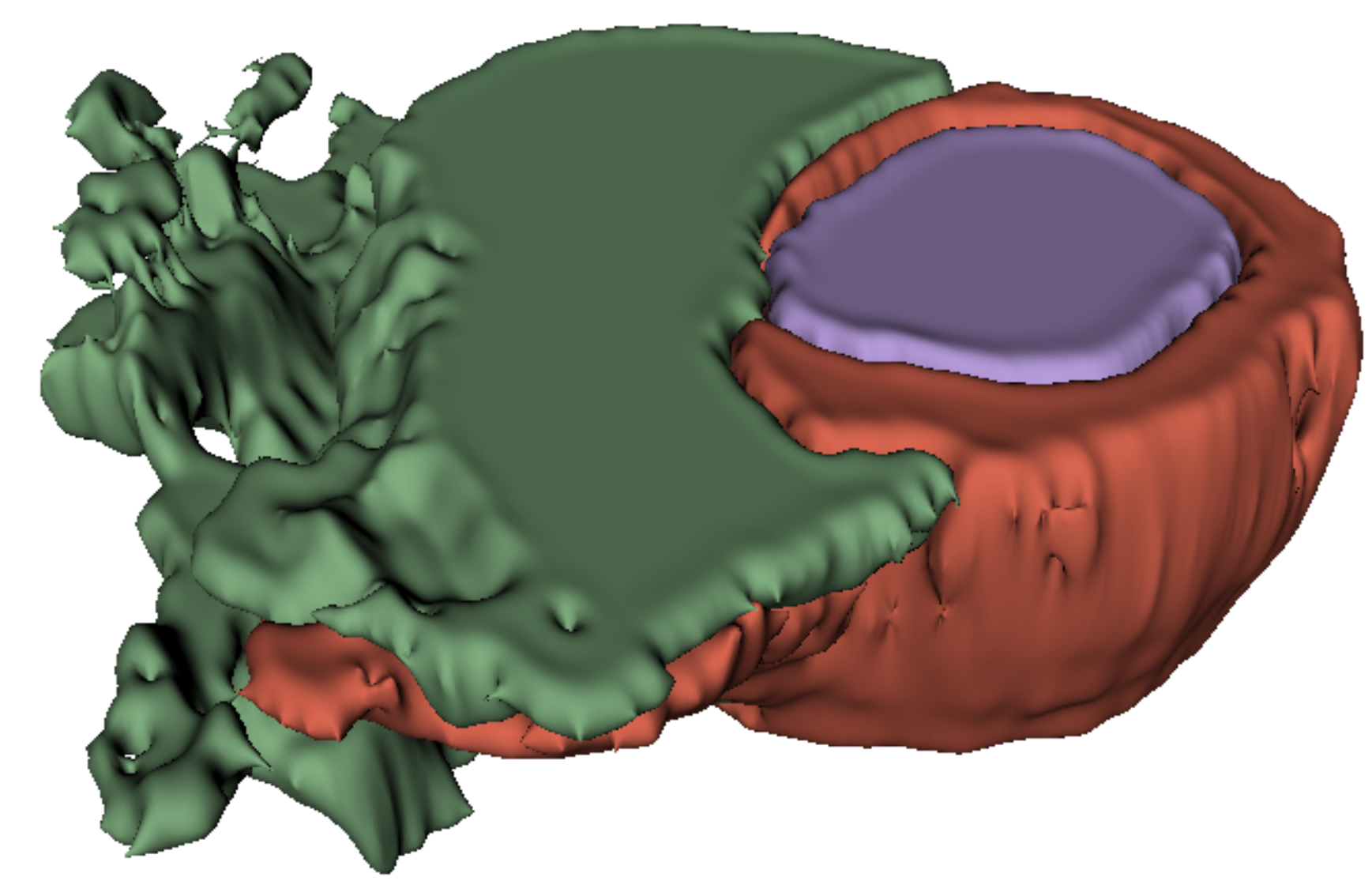}
         \label{}
     \end{subfigure}
     \begin{subfigure}[b]{0.13\textwidth}
         \centering
         \includegraphics[width=\textwidth, height=0.07\textheight]{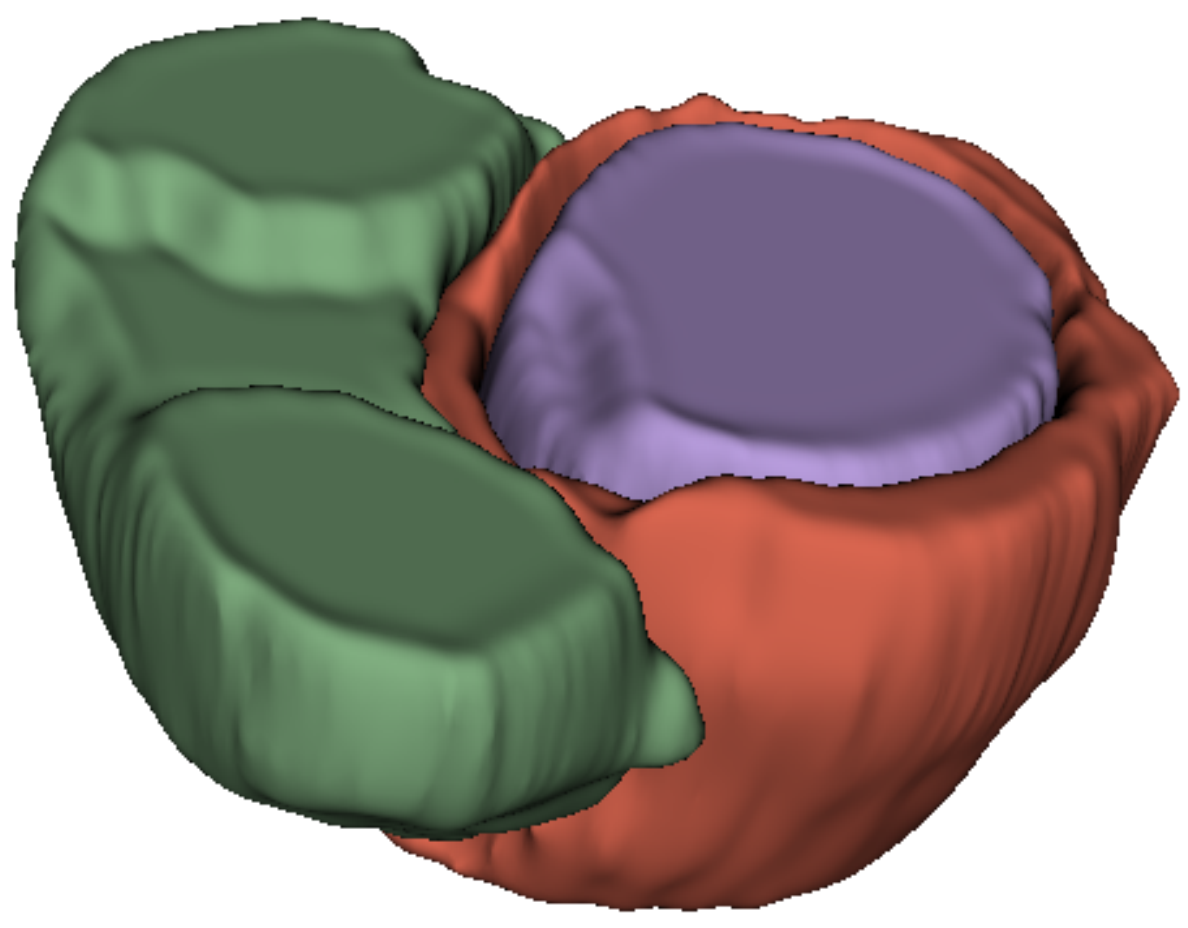}
         \label{}
     \end{subfigure}\\[-2.7ex]

     \begin{subfigure}[b]{0.13\textwidth}
         \centering
         \includegraphics[width=\textwidth, height=0.07\textheight]{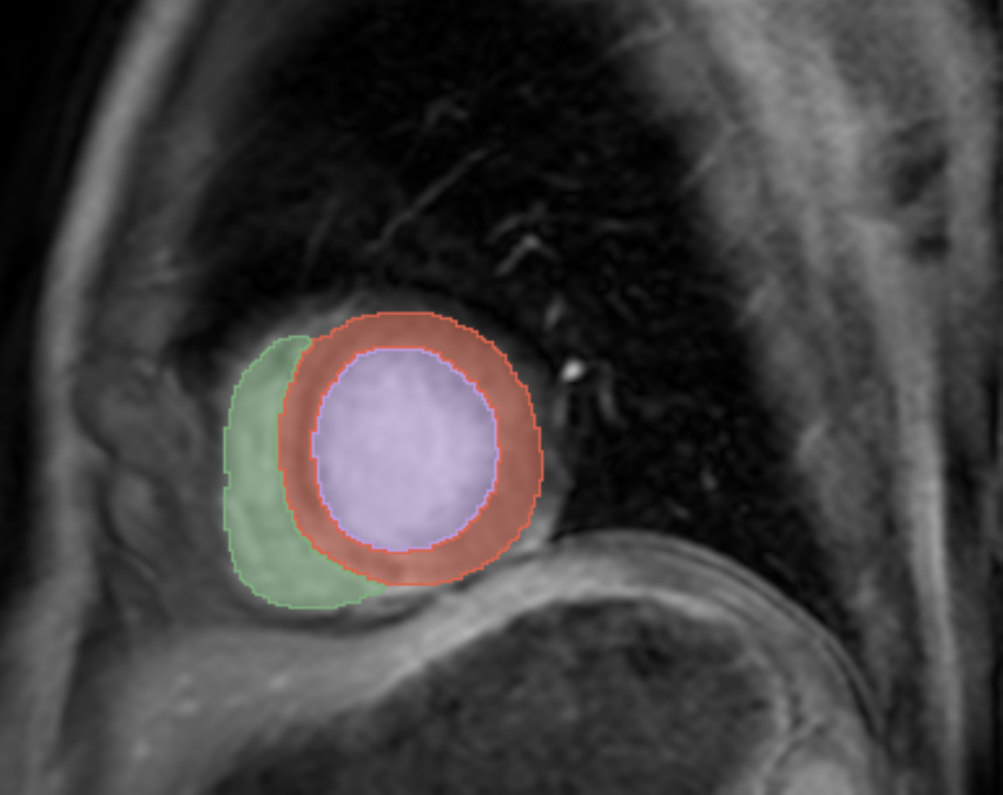}
         \label{}
     \end{subfigure}
     \begin{subfigure}[b]{0.13\textwidth}
         \centering
         \includegraphics[width=\textwidth, height=0.07\textheight]{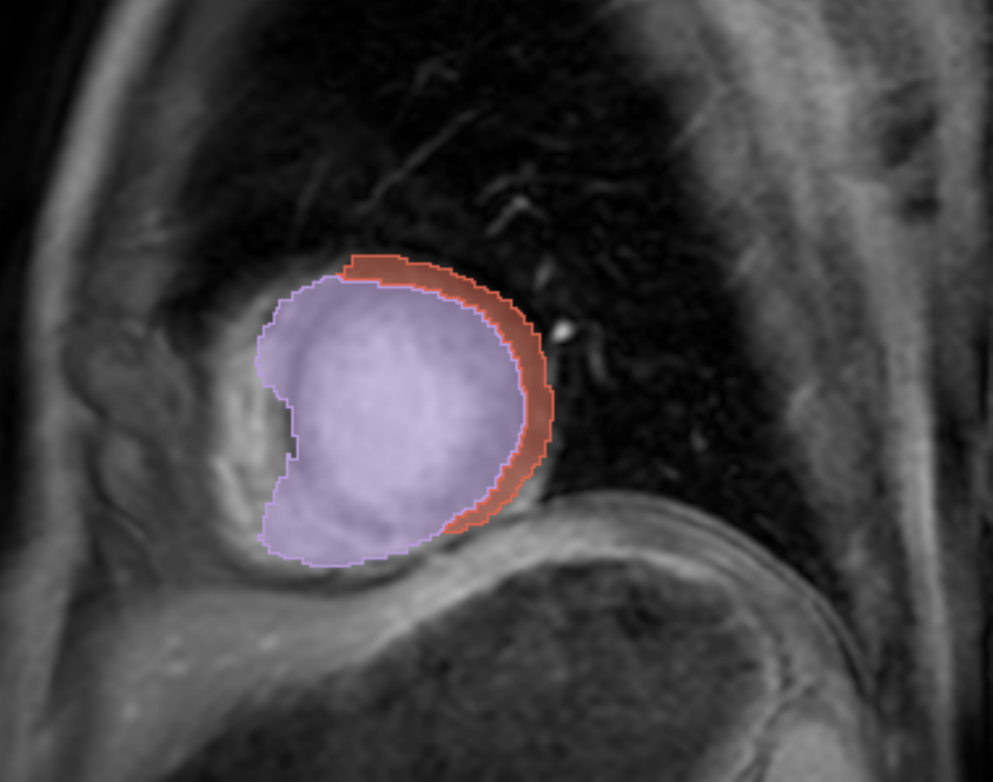}
         \label{}
     \end{subfigure}
      \begin{subfigure}[b]{0.13\textwidth}
         \centering
         \includegraphics[width=\textwidth, height=0.07\textheight]{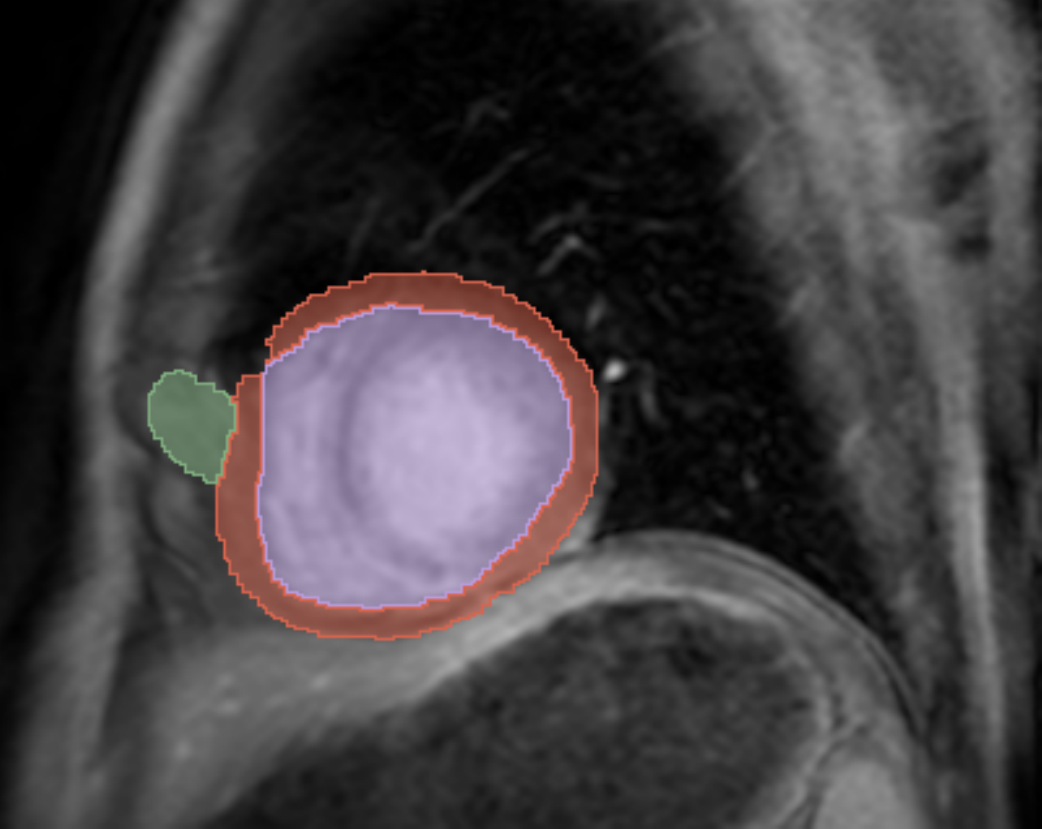}
         \label{}
     \end{subfigure}
    \begin{subfigure}[b]{0.13\textwidth}
         \centering
         \includegraphics[width=\textwidth, height=0.07\textheight]{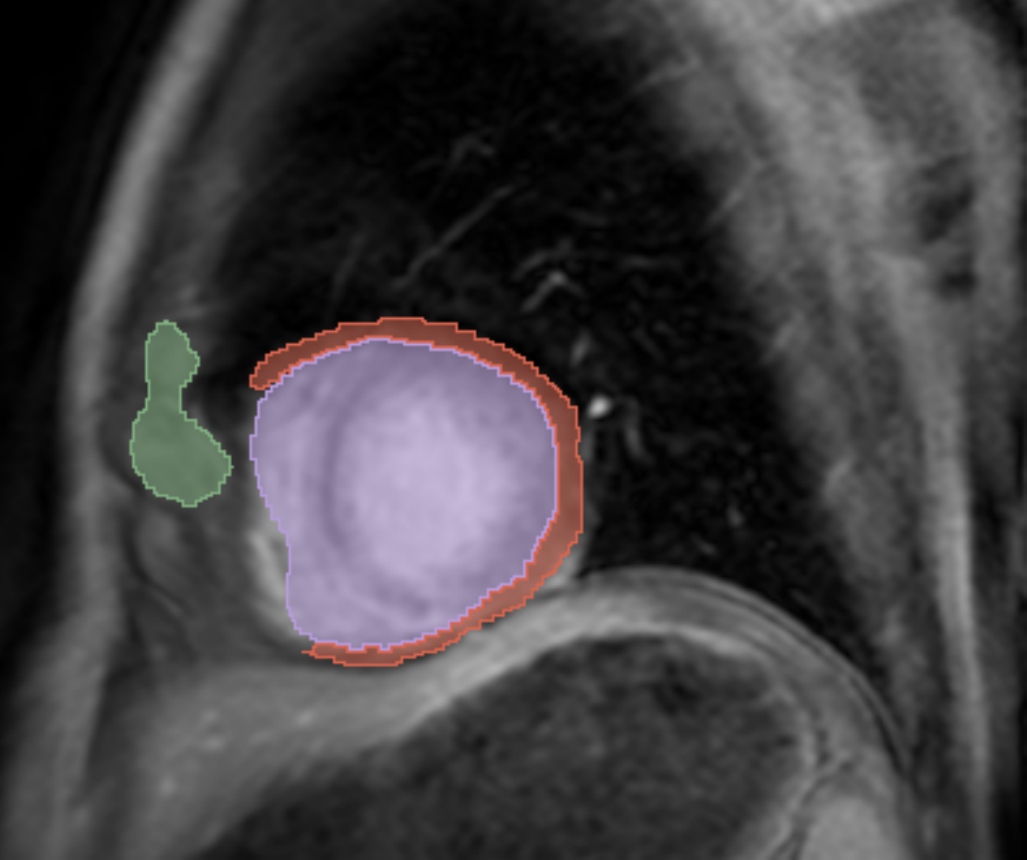}
         \label{}
     \end{subfigure}
    \begin{subfigure}[b]{0.13\textwidth}
         \centering
         \includegraphics[width=\textwidth, height=0.07\textheight]{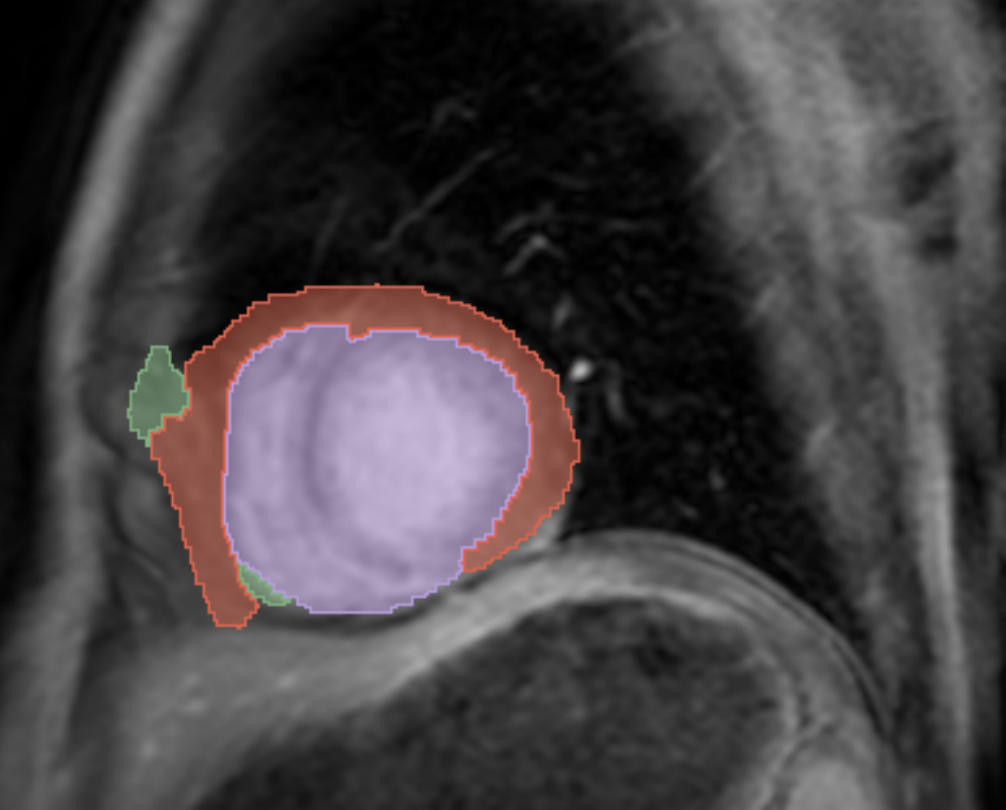}
         \label{}
     \end{subfigure}
     \begin{subfigure}[b]{0.13\textwidth}
         \centering
         \includegraphics[width=\textwidth, height=0.07\textheight]{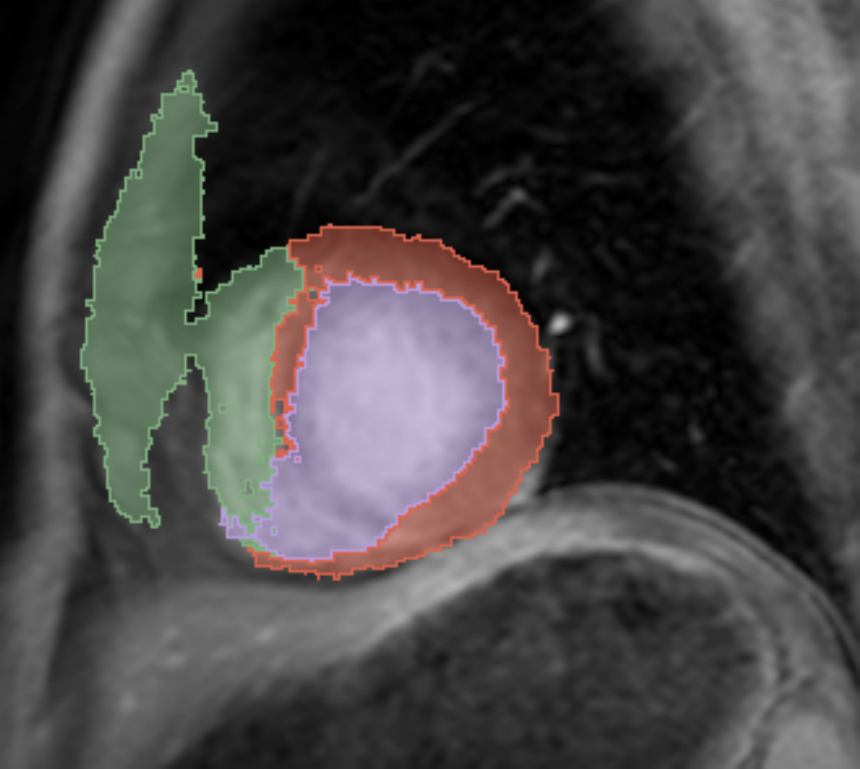}
         \label{}
     \end{subfigure}
     \begin{subfigure}[b]{0.13\textwidth}
         \centering
         \includegraphics[width=\textwidth, height=0.07\textheight]{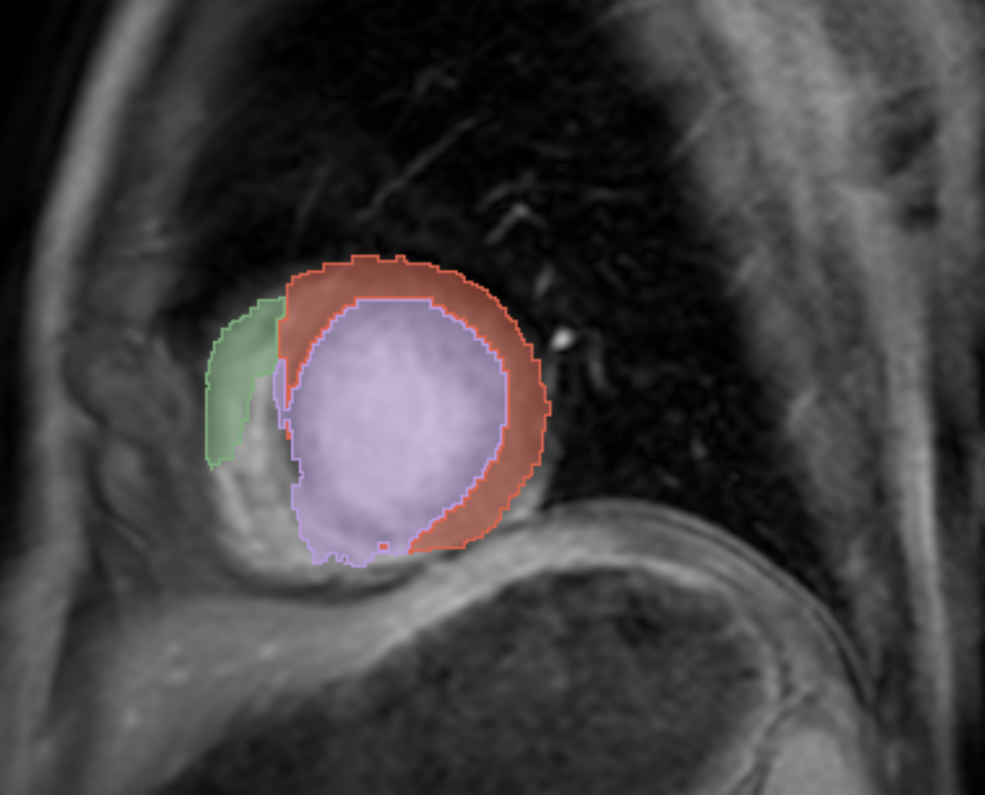}
         \label{}
     \end{subfigure} \\[-2.7ex]

     \begin{subfigure}[b]{0.13\textwidth}
         \centering
         \includegraphics[width=\textwidth, height=0.07\textheight]{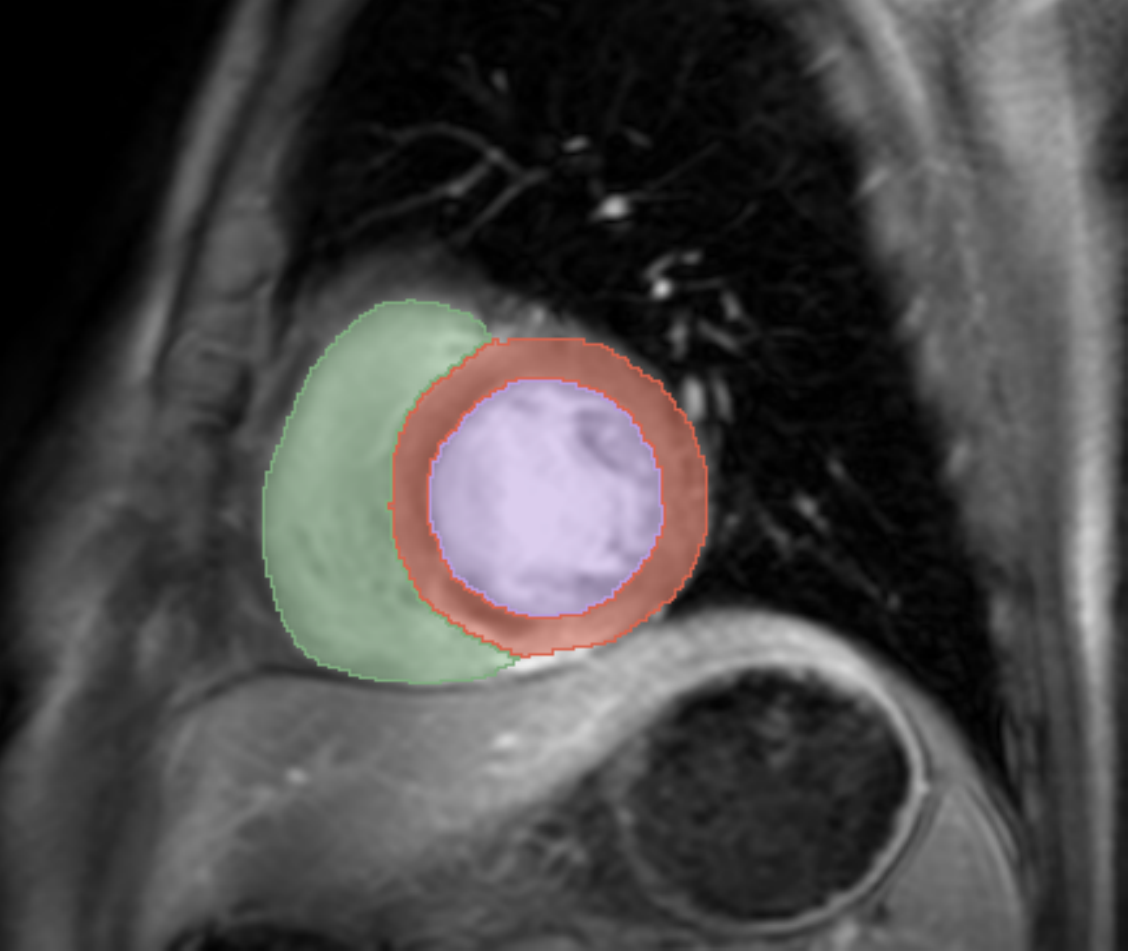}
         \label{}
     \end{subfigure}
     \begin{subfigure}[b]{0.13\textwidth}
         \centering
         \includegraphics[width=\textwidth, height=0.07\textheight]{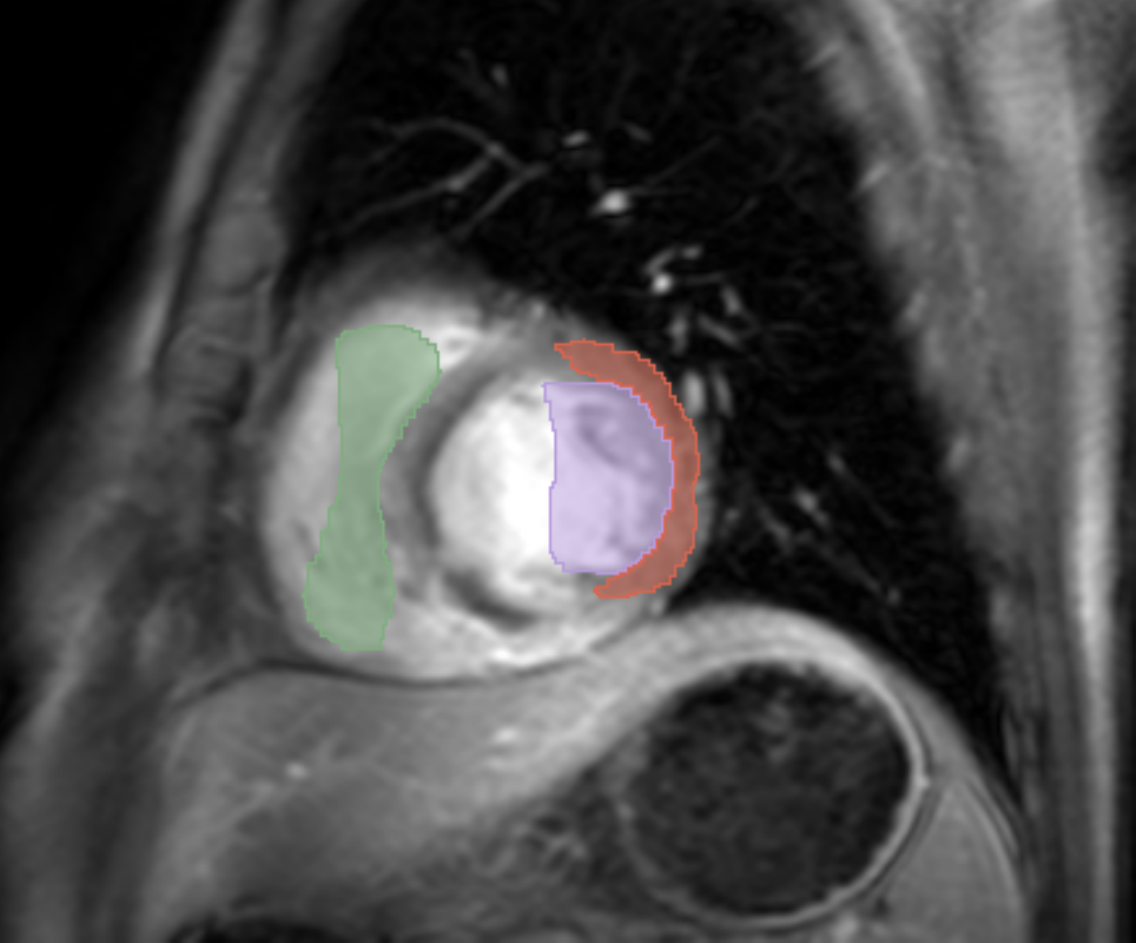}
         \label{}
     \end{subfigure}
      \begin{subfigure}[b]{0.13\textwidth}
         \centering
         \includegraphics[width=\textwidth, height=0.07\textheight]{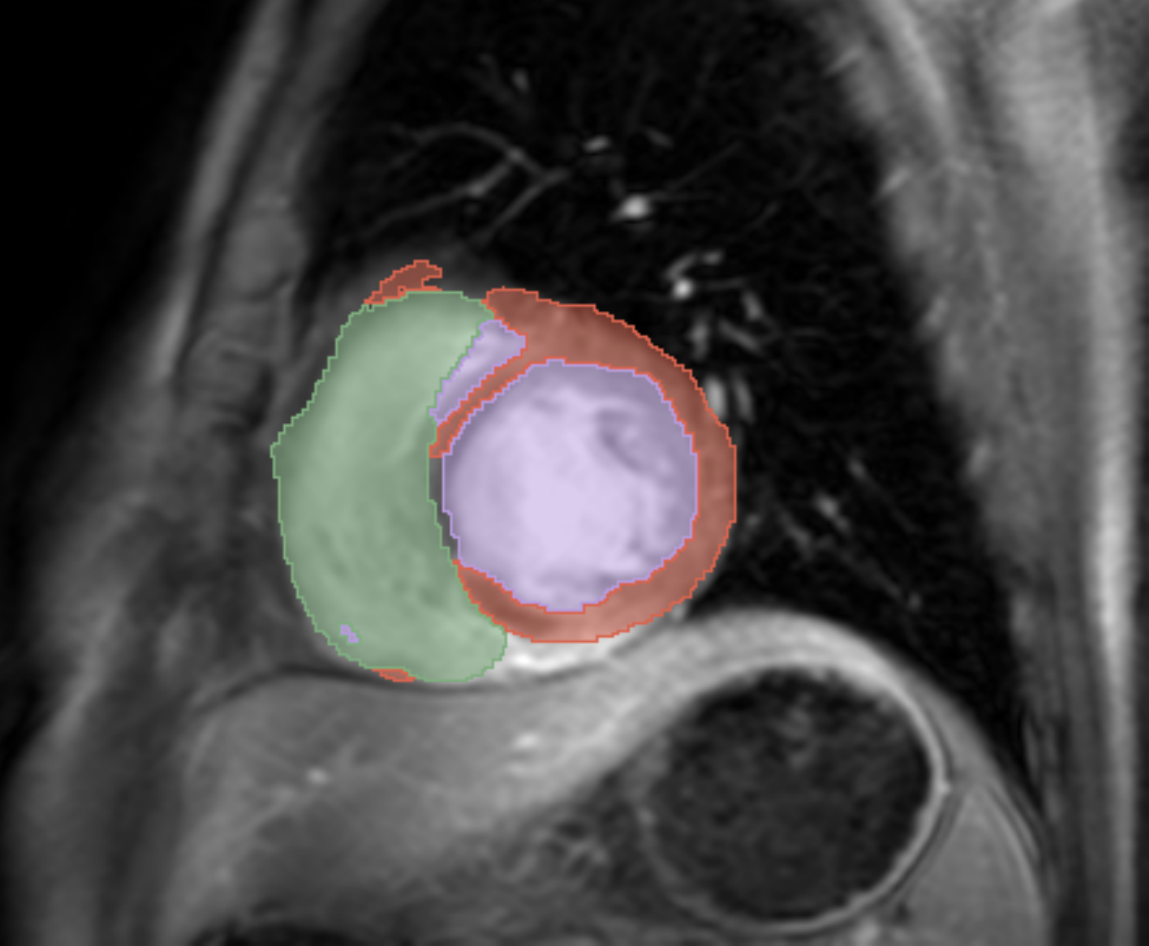}
         \label{}
     \end{subfigure}
    \begin{subfigure}[b]{0.13\textwidth}
         \centering
         \includegraphics[width=\textwidth, height=0.07\textheight]{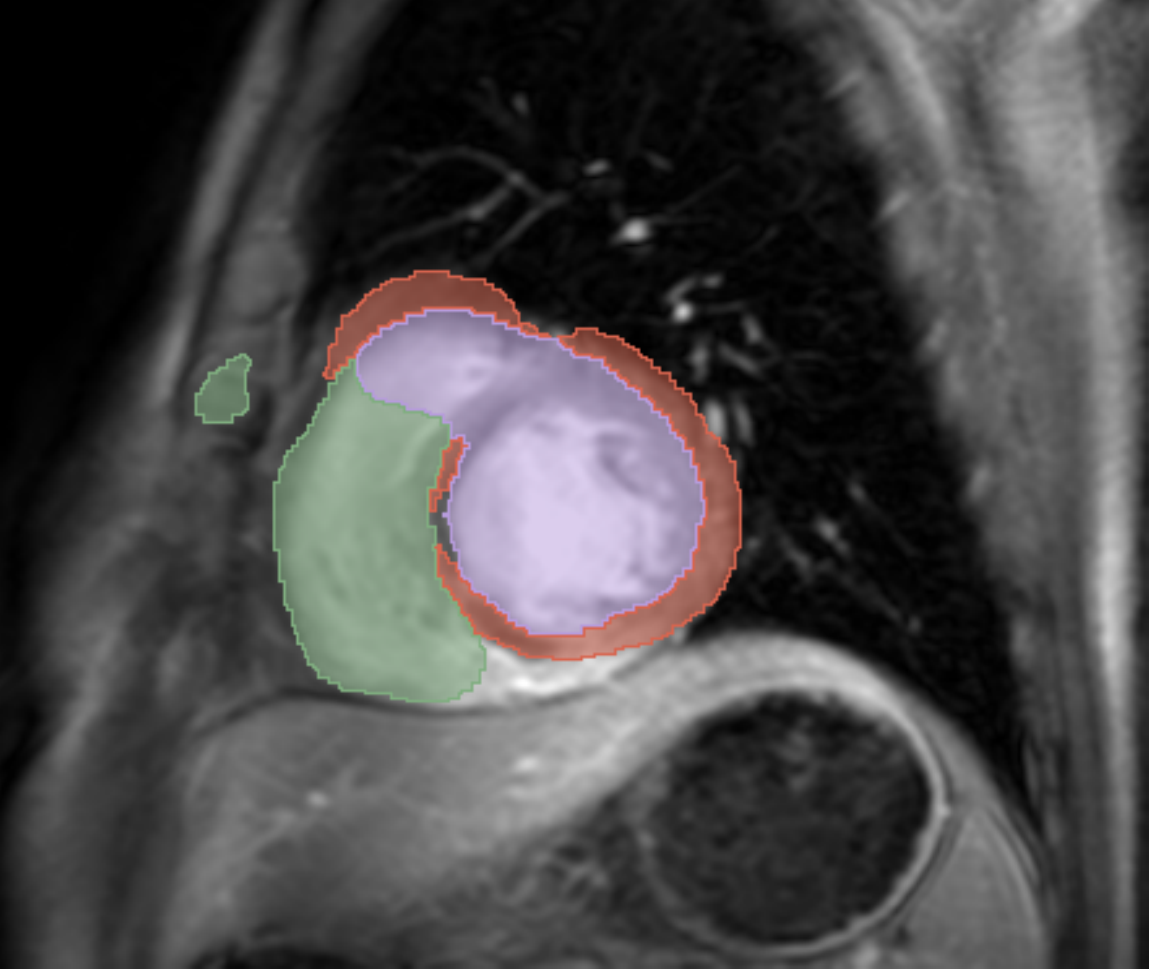}
         \label{}
     \end{subfigure}
    \begin{subfigure}[b]{0.13\textwidth}
         \centering
         \includegraphics[width=\textwidth, height=0.07\textheight]{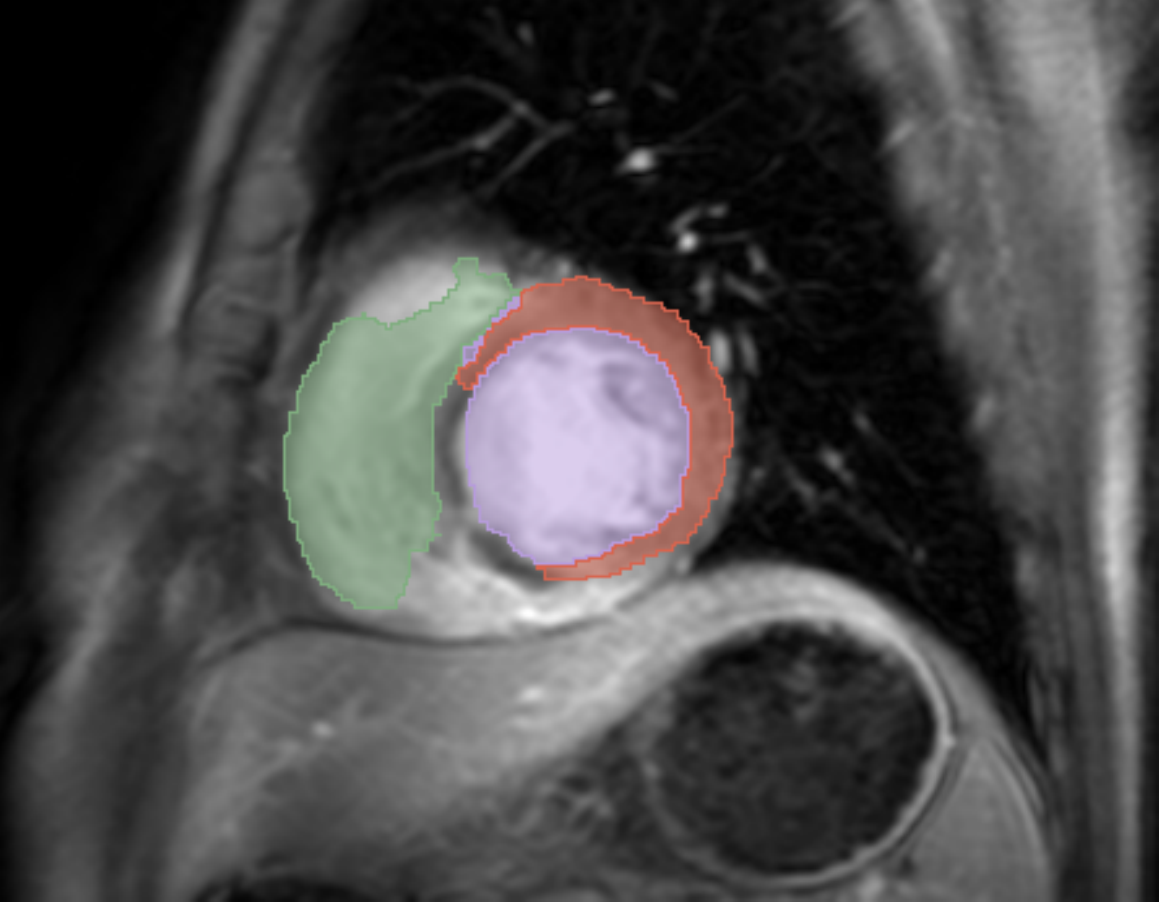}
         \label{}
     \end{subfigure}
     \begin{subfigure}[b]{0.13\textwidth}
         \centering
         \includegraphics[width=\textwidth, height=0.07\textheight]{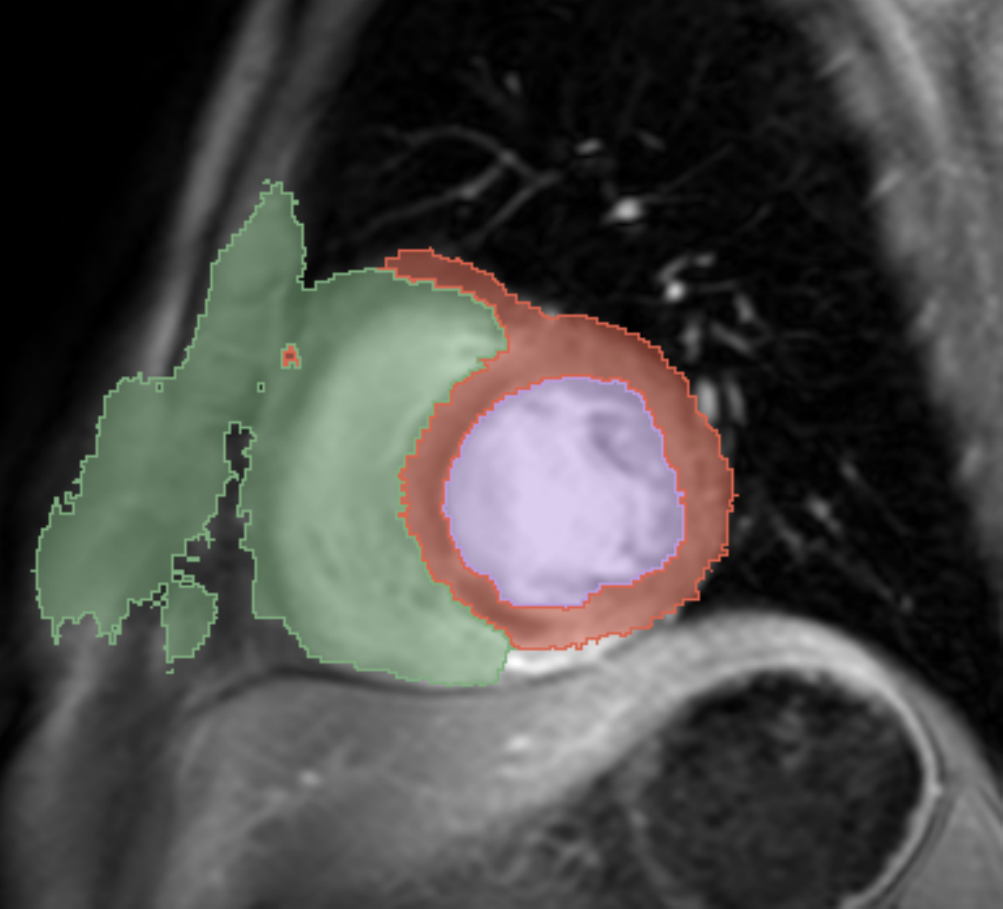}
         \label{}
     \end{subfigure}
     \begin{subfigure}[b]{0.13\textwidth}
         \centering
         \includegraphics[width=\textwidth, height=0.07\textheight]{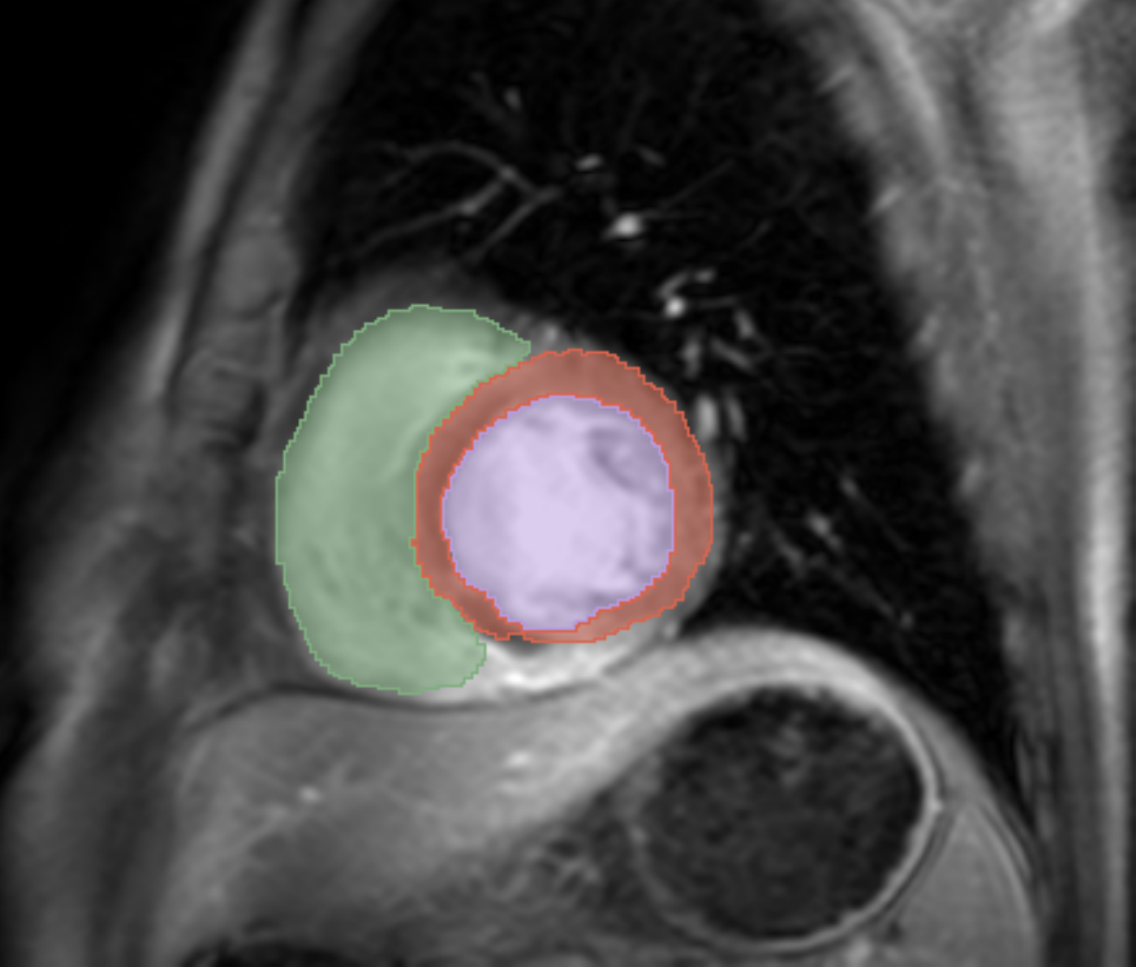}
         \label{}
     \end{subfigure} \\[-2.8ex]

     \begin{subfigure}[b]{0.13\textwidth}
         \centering
         \includegraphics[width=\textwidth, height=0.07\textheight]{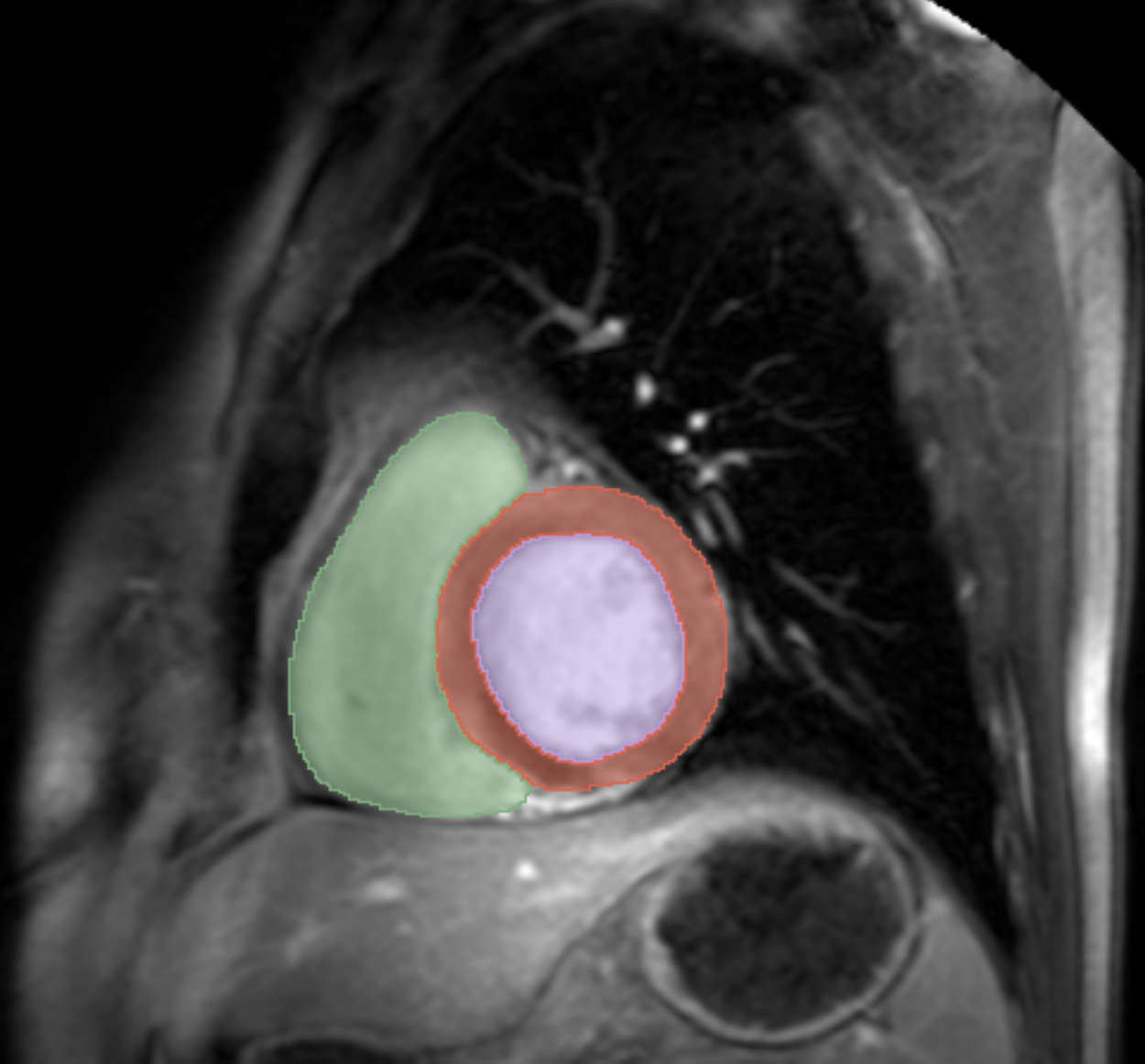}
        \caption{GT}
     \end{subfigure}
     \begin{subfigure}[b]{0.13\textwidth}
         \centering
         \includegraphics[width=\textwidth, height=0.07\textheight]{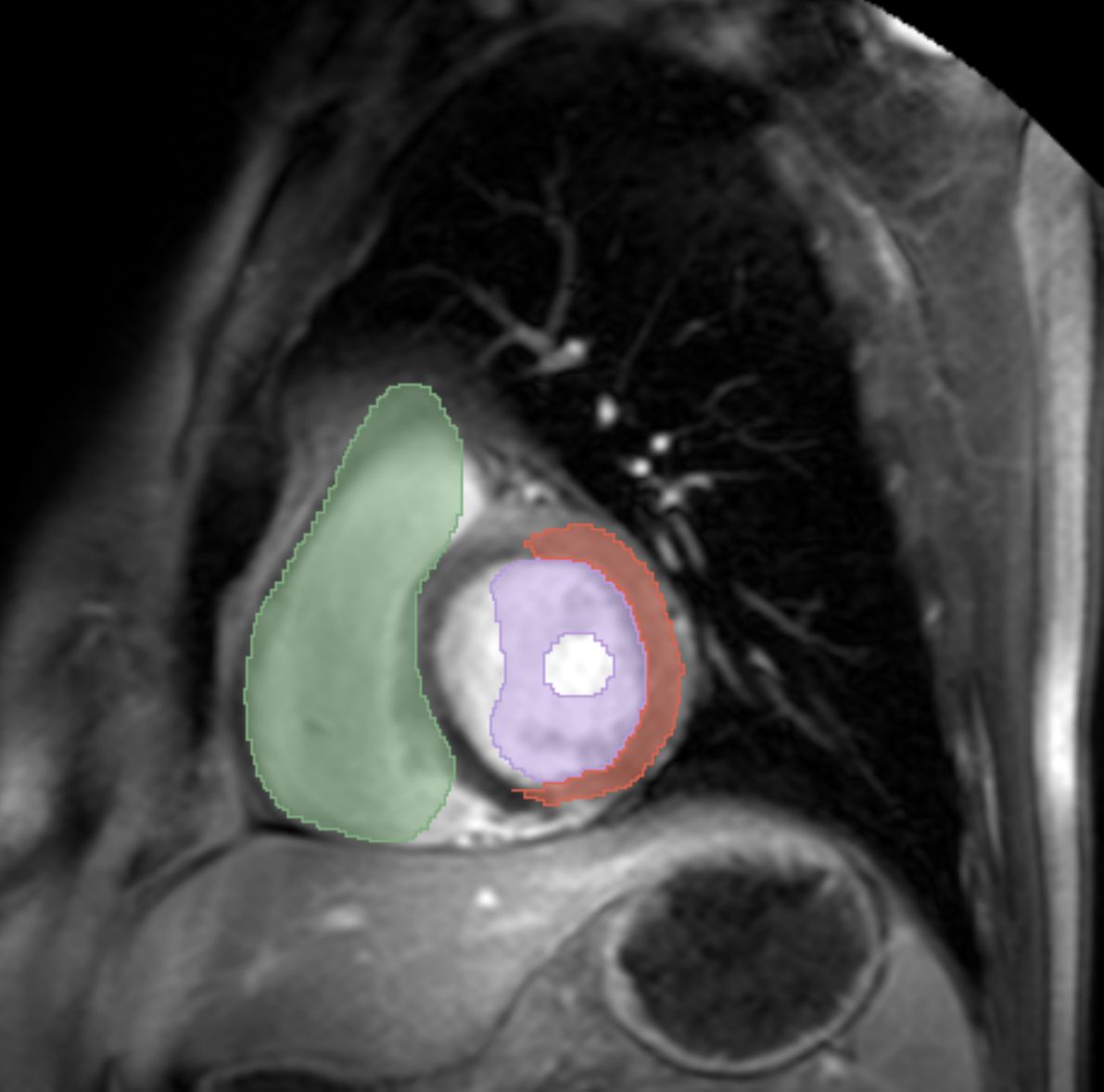}
        \caption{w/oUDA}
     \end{subfigure}
     \begin{subfigure}[b]{0.13\textwidth}
         \centering
         \includegraphics[width=\textwidth, height=0.07\textheight]{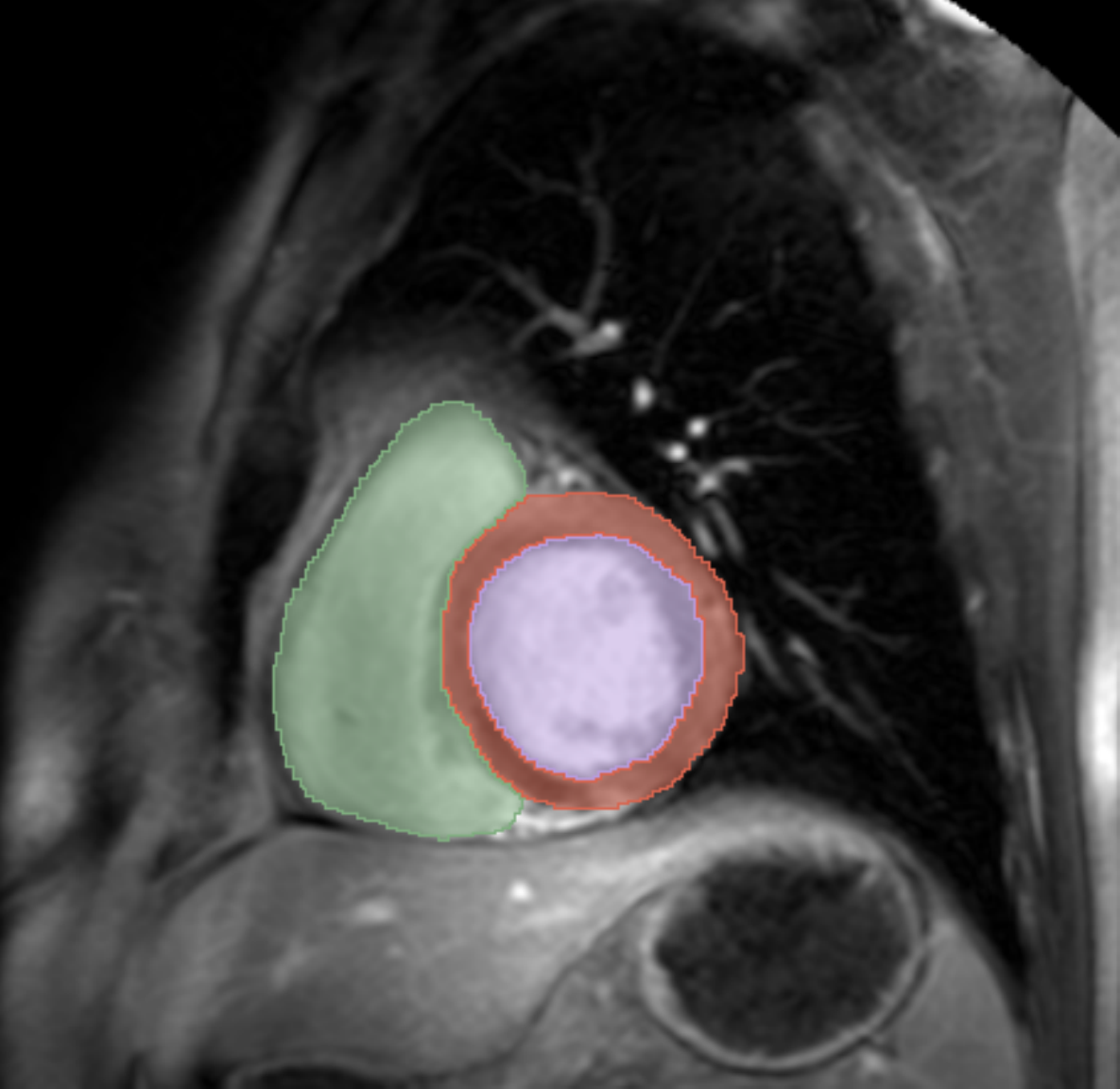}
        \caption{AdaptSeg}
     \end{subfigure}
    \begin{subfigure}[b]{0.13\textwidth}
         \centering
         \includegraphics[width=\textwidth, height=0.07\textheight]{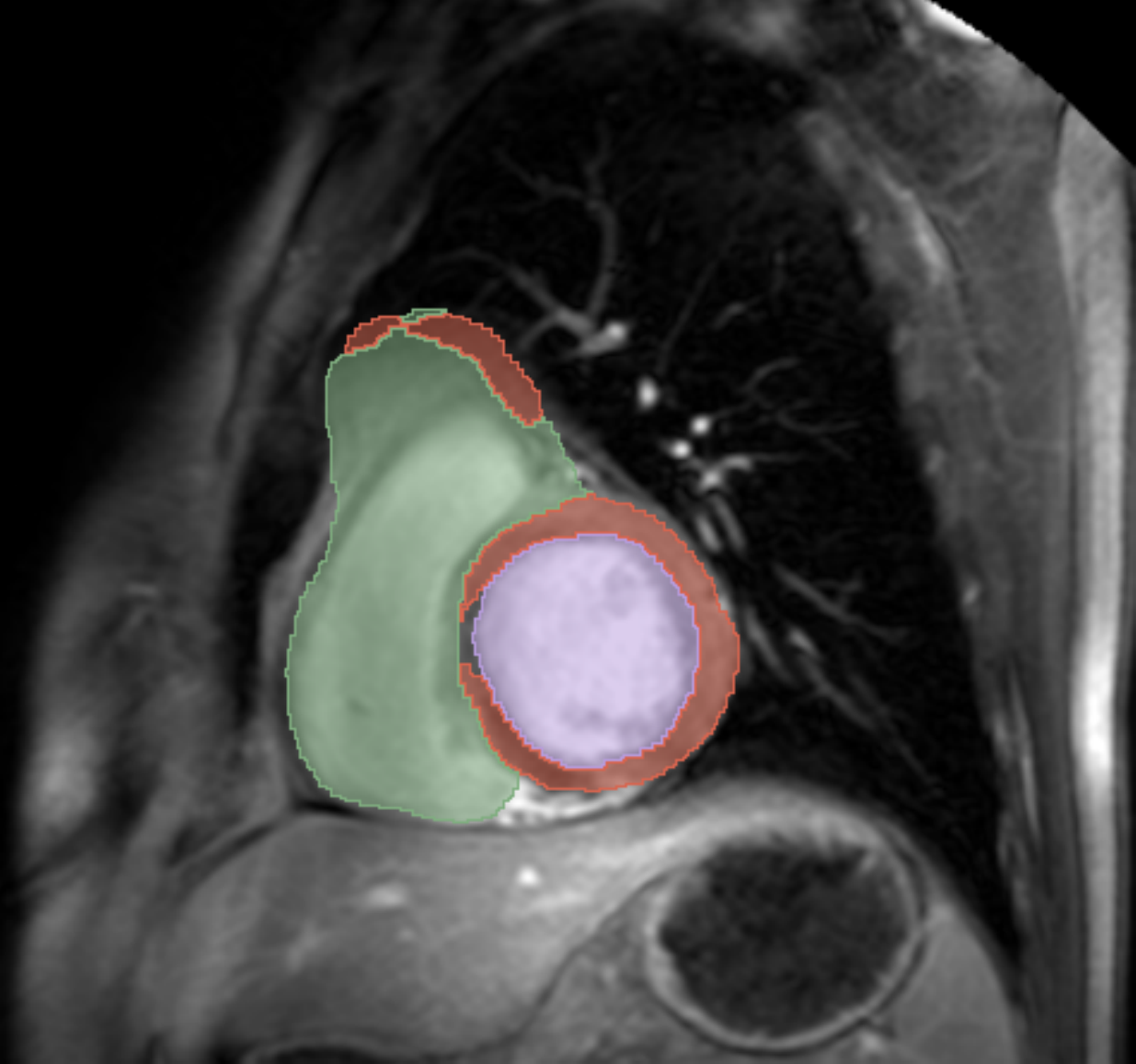}
        \caption{AdvEnt}
     \end{subfigure}
    \begin{subfigure}[b]{0.13\textwidth}
         \centering
         \includegraphics[width=\textwidth, height=0.07\textheight]{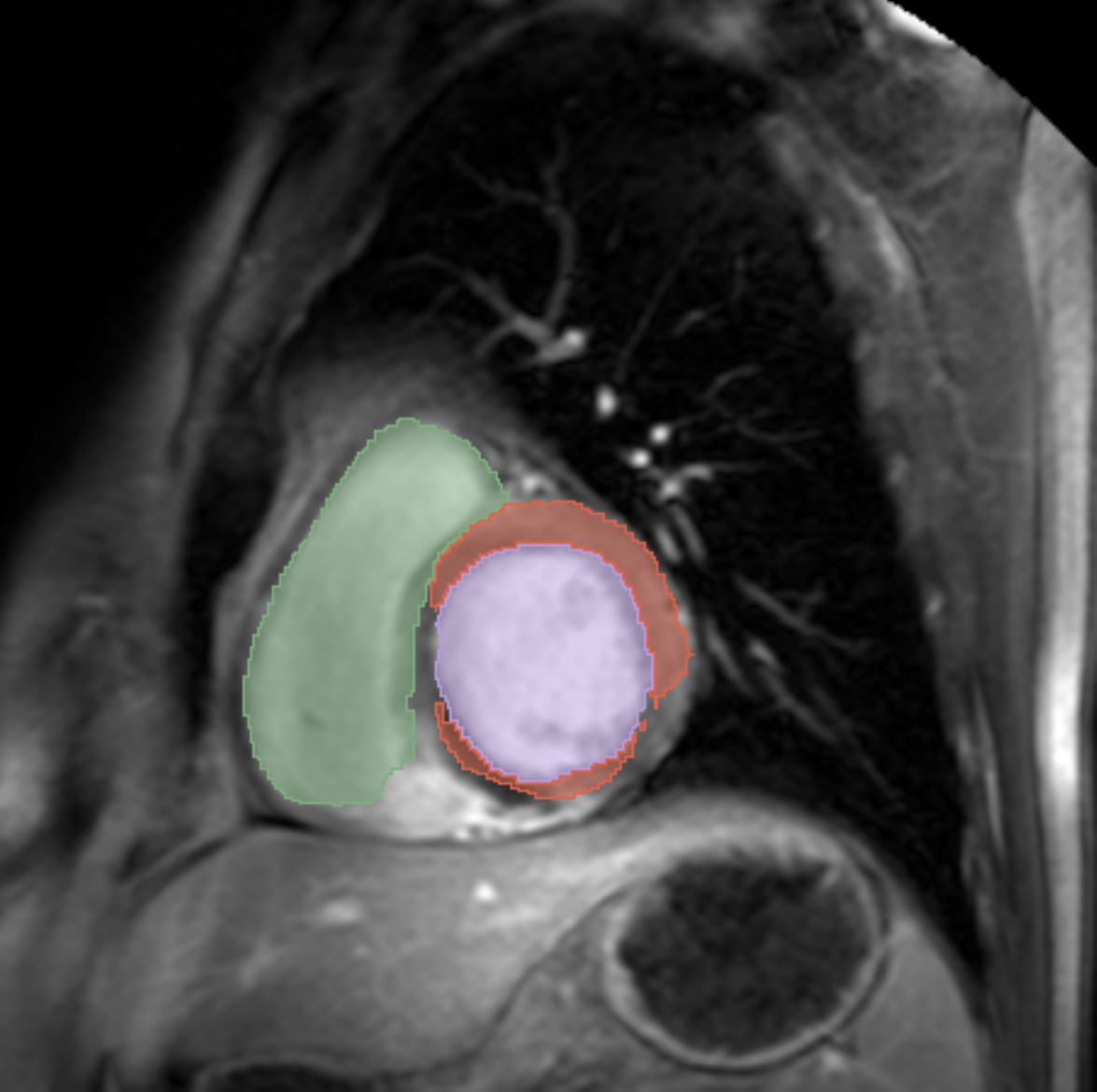}
        \caption{FUDA}
     \end{subfigure}
    \begin{subfigure}[b]{0.13\textwidth}
         \centering
         \includegraphics[width=\textwidth, height=0.07\textheight]{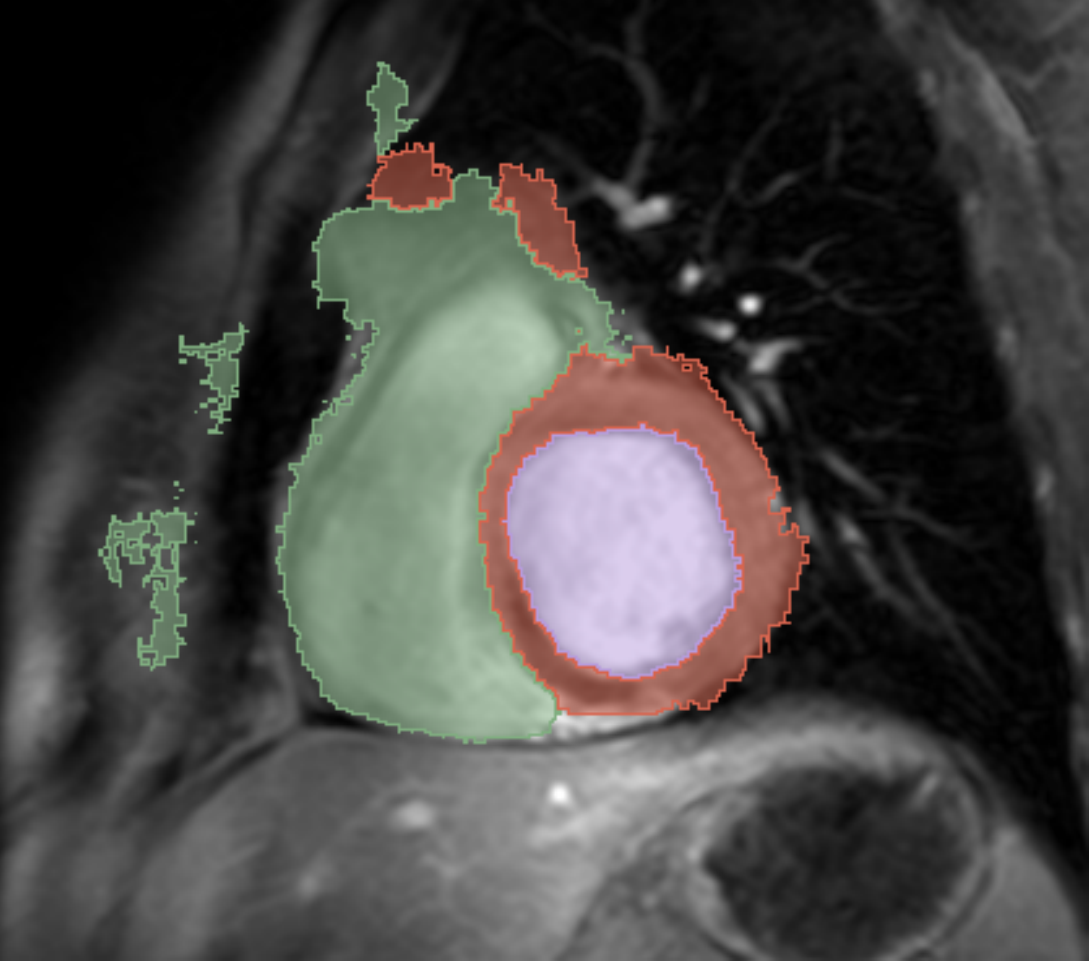}
        \caption{BlockCL}
     \end{subfigure}
     \begin{subfigure}[b]{0.13\textwidth}
         \centering
         \includegraphics[width=\textwidth, height=0.07\textheight]{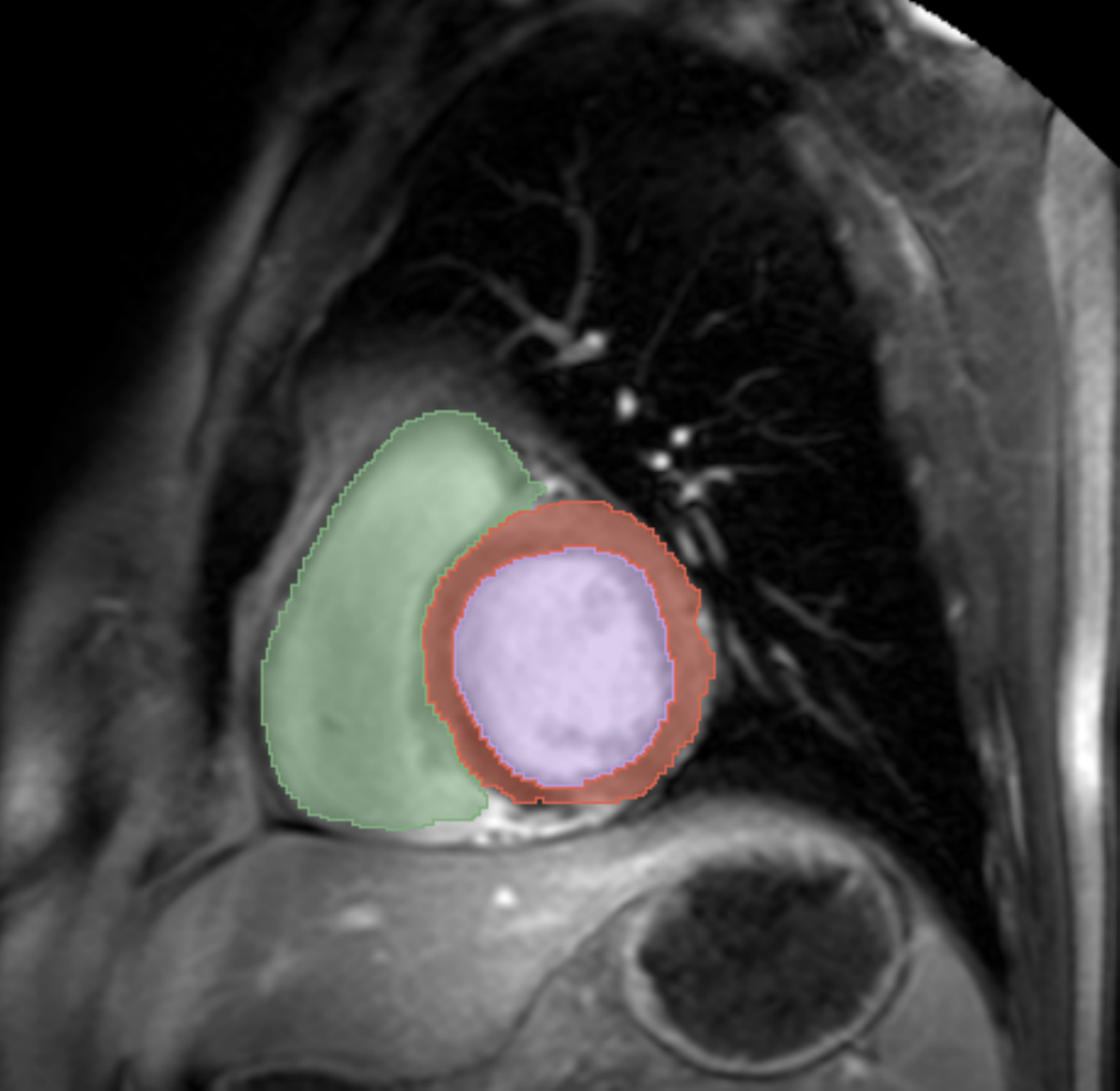}
        \caption{Ours}
     \end{subfigure} 
    \caption{Segmentation results under fewshot for one LGE-CMR in 3D and 3 slices of the sequence in 2D from the test set using different methods. It shows: (a) the ground-truth segmentation, (b) the prediction of W/o UDA, (c) AdaptSeg \cite{3260-08}, (d) AdvEnt \cite{3260-03}, (e) FUDA \cite{gu2022few}, (f) BlockCL \cite{Hu2021}, and (g) ConFUDA (Ours). The LV is shown in purple, MYO in red, and RV in green color.}
    \label{fig:three graphs}
\end{figure}
\noindent\textbf{Implementation details.}
DR-UNet \cite{3260-09} is used as the segmentation network. We perform 5-fold cross-validation to have a robust evaluation for all the methods. bSSFP and T2 sequences are used to pre-train the RAIN module. The segmentation network is then trained with bSSFP as source data and LGE as target data. Pre-training of RAIN does not require target LGE images, so it can be trained before LGE acquisition. For each experiment, we first warmed up the segmentation module for 200 epochs with the RAIN module. Style updating ($\epsilon$) was switched off at this stage due to unreliable target predictions. After that, the models were trained for another 200 epochs with the $\epsilon$ turned on. The SGD optimizer is used for all training. The learning rate is set to 5e-4 for warm-up and 2.5e-4 for the rest of the training with a polynomial decay. $\lambda_{contrast}$ and $\lambda_{CNR}$ are set to 1 and 0.5, respectively. The batch size of source images was 32, and that of target images was one, consistent with the oneshot setting. 

\subsection{Performance Evaluation and Discussion}
Qualitative comparison of the proposed ConFUDA with other well-known UDA methods are shown in \cref{fig:three graphs}. All the comparison methods are trained with bSSFP and T2 and tested on LGE sequences. We can observe that the proposed ConFUDA is able to generate more precise and compact cardiac structures among all the compared methods. Quantitative comparison is shown in \cref{tab:comparision}. Compared with the baseline, we achieved more than 0.3 of DC for all the experimental settings. We surpass the famous UDA methods AdaptSeg and AdvEnt in both DC and HD95. BlockCL has a high HD95 which follows the visualization in \cref{fig:three graphs}. We reduce the HD95 of BlockCL by 1.3$\sim$7.2 which results in fewer outliers.

\begin{figure}[!tb]
     \centering
    \begin{subfigure}[b]{0.16\textwidth}
         \centering
         \includegraphics[width=\textwidth, height=0.08\textheight]{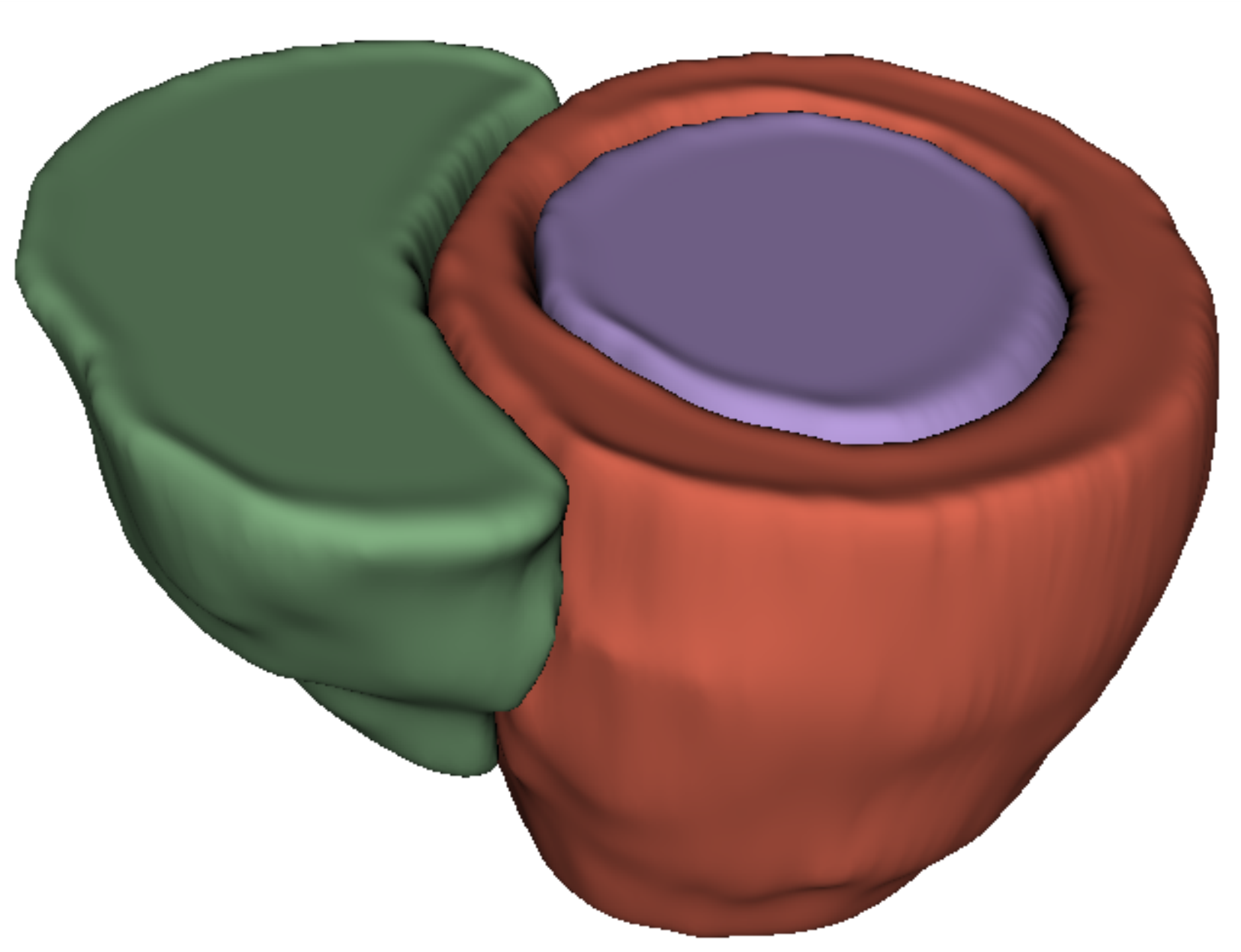}
         \label{}
     \end{subfigure}
    \begin{subfigure}[b]{0.16\textwidth}
         \centering
         \includegraphics[width=\textwidth, height=0.08\textheight]{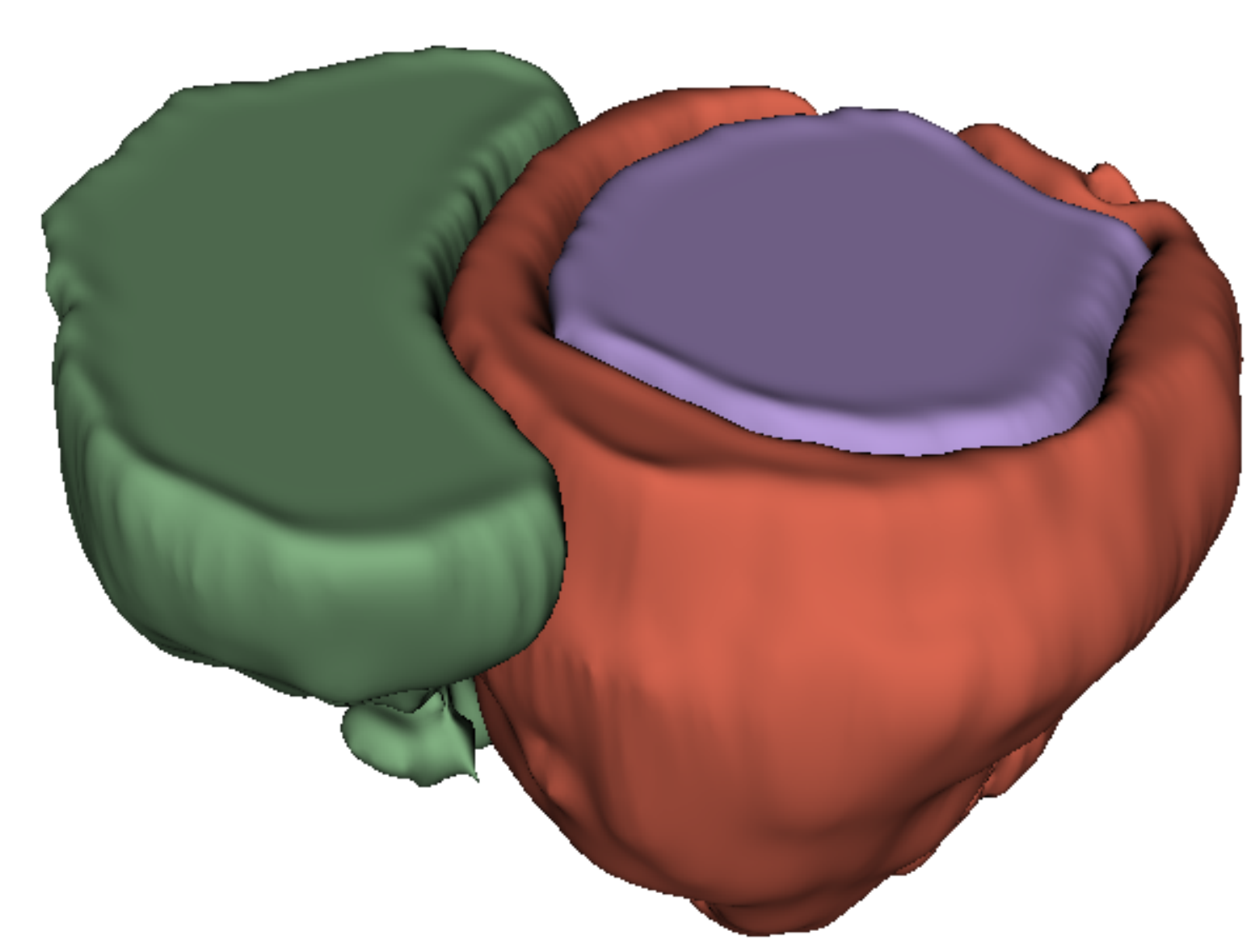}
         \label{}
     \end{subfigure}
    \begin{subfigure}[b]{0.16\textwidth}
         \centering
         \includegraphics[width=\textwidth, height=0.08\textheight]{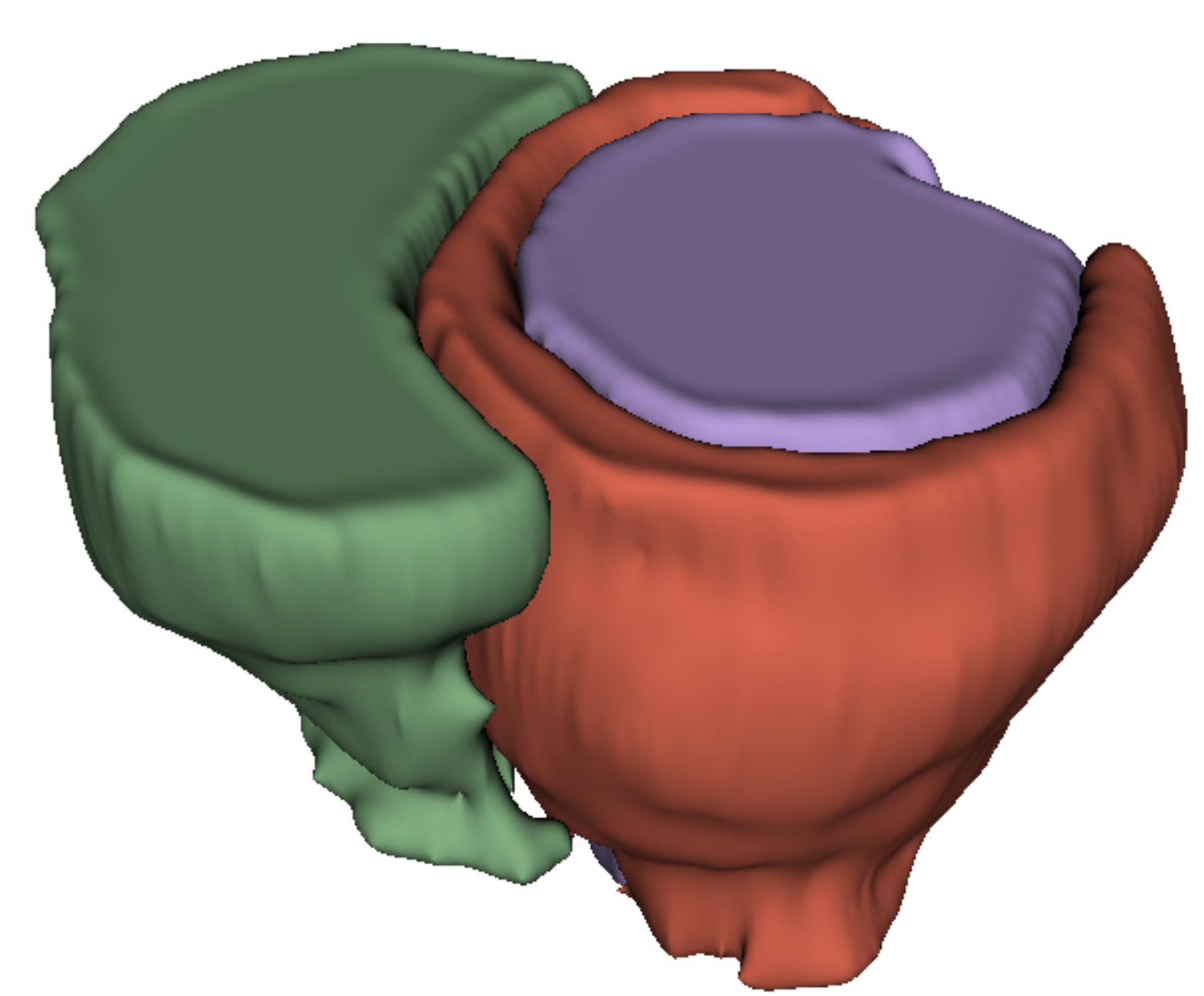}
         \label{}
     \end{subfigure}
    \begin{subfigure}[b]{0.16\textwidth}
         \centering
         \includegraphics[width=\textwidth, height=0.08\textheight]{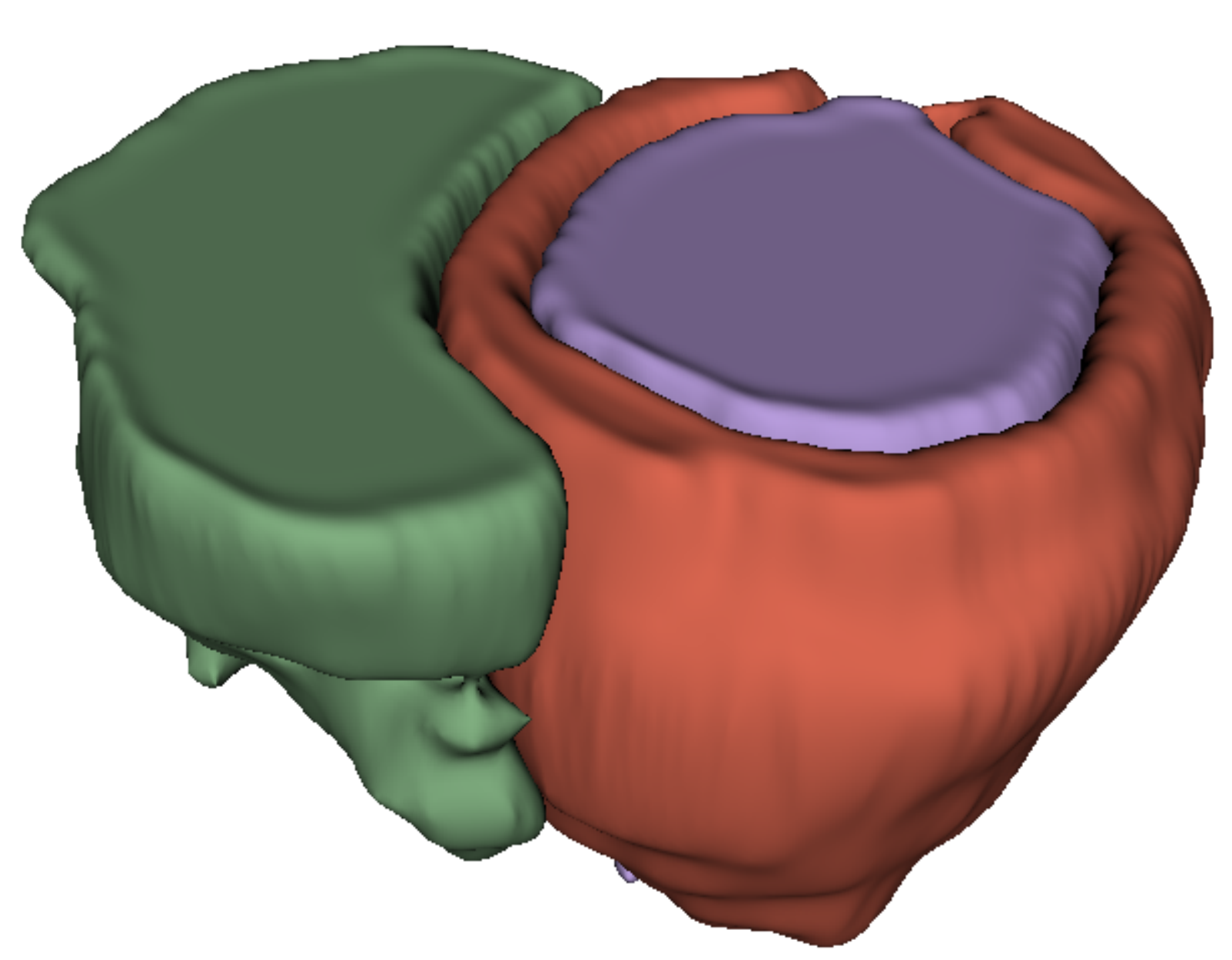}
         \label{}
     \end{subfigure}
     \begin{subfigure}[b]{0.16\textwidth}
         \centering
         \includegraphics[width=\textwidth, height=0.08\textheight]{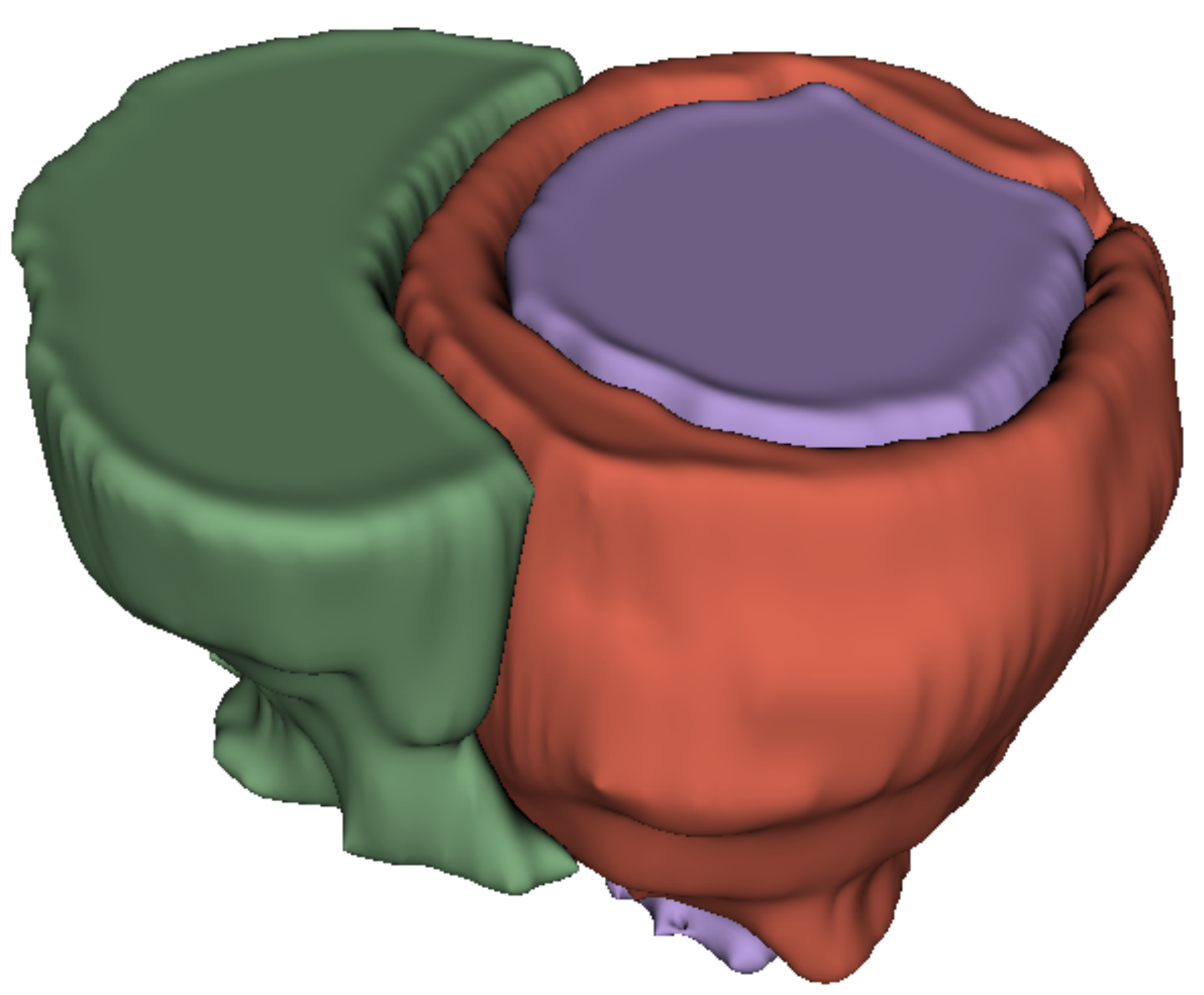}
         \label{}
     \end{subfigure}\\[-2.7ex]
     
    \begin{subfigure}[b]{0.16\textwidth}
         \centering
         \includegraphics[width=\textwidth, height=0.08\textheight]{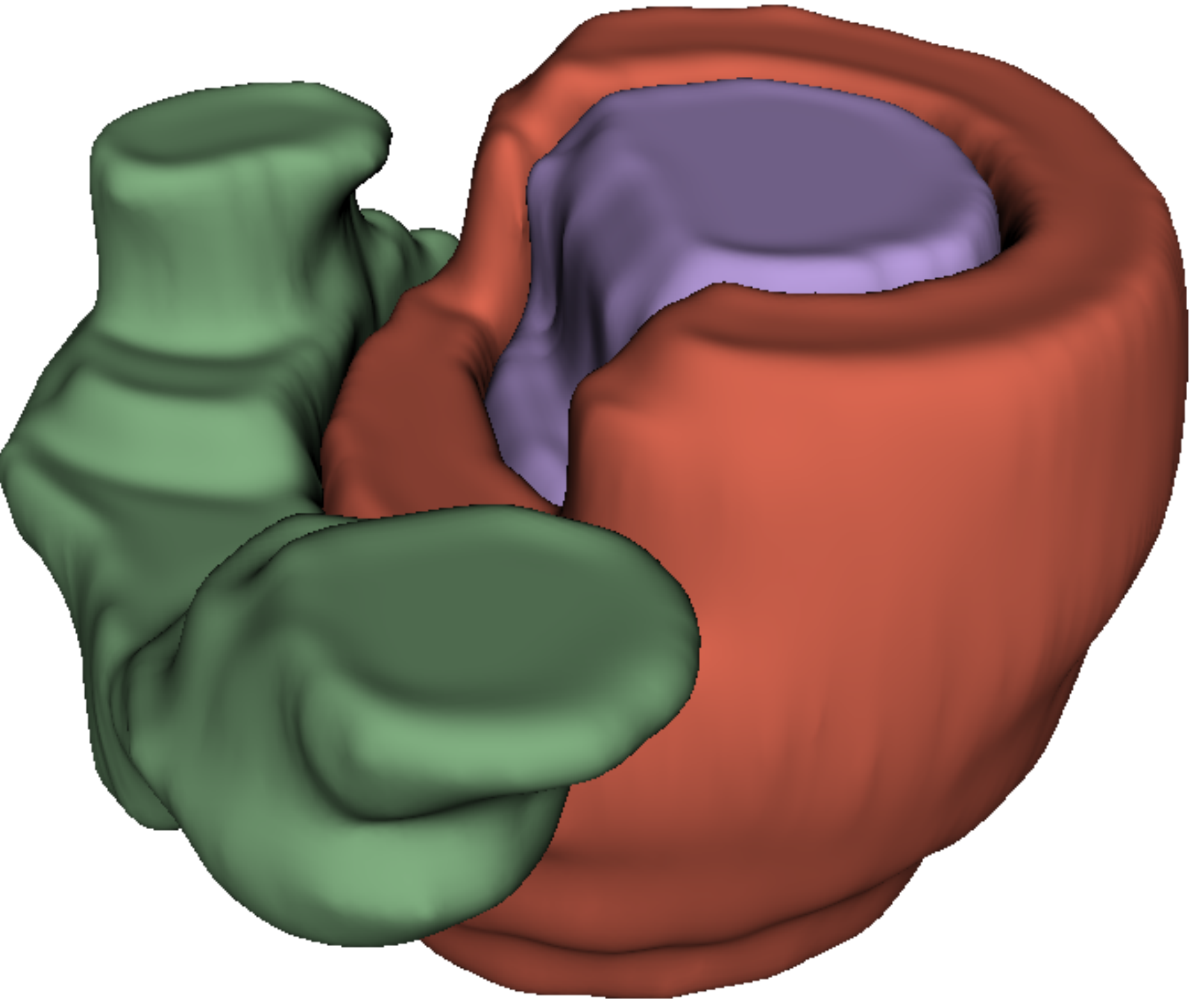}
        \caption{}
     \end{subfigure}
    \begin{subfigure}[b]{0.16\textwidth}
         \centering
         \includegraphics[width=\textwidth, height=0.08\textheight]{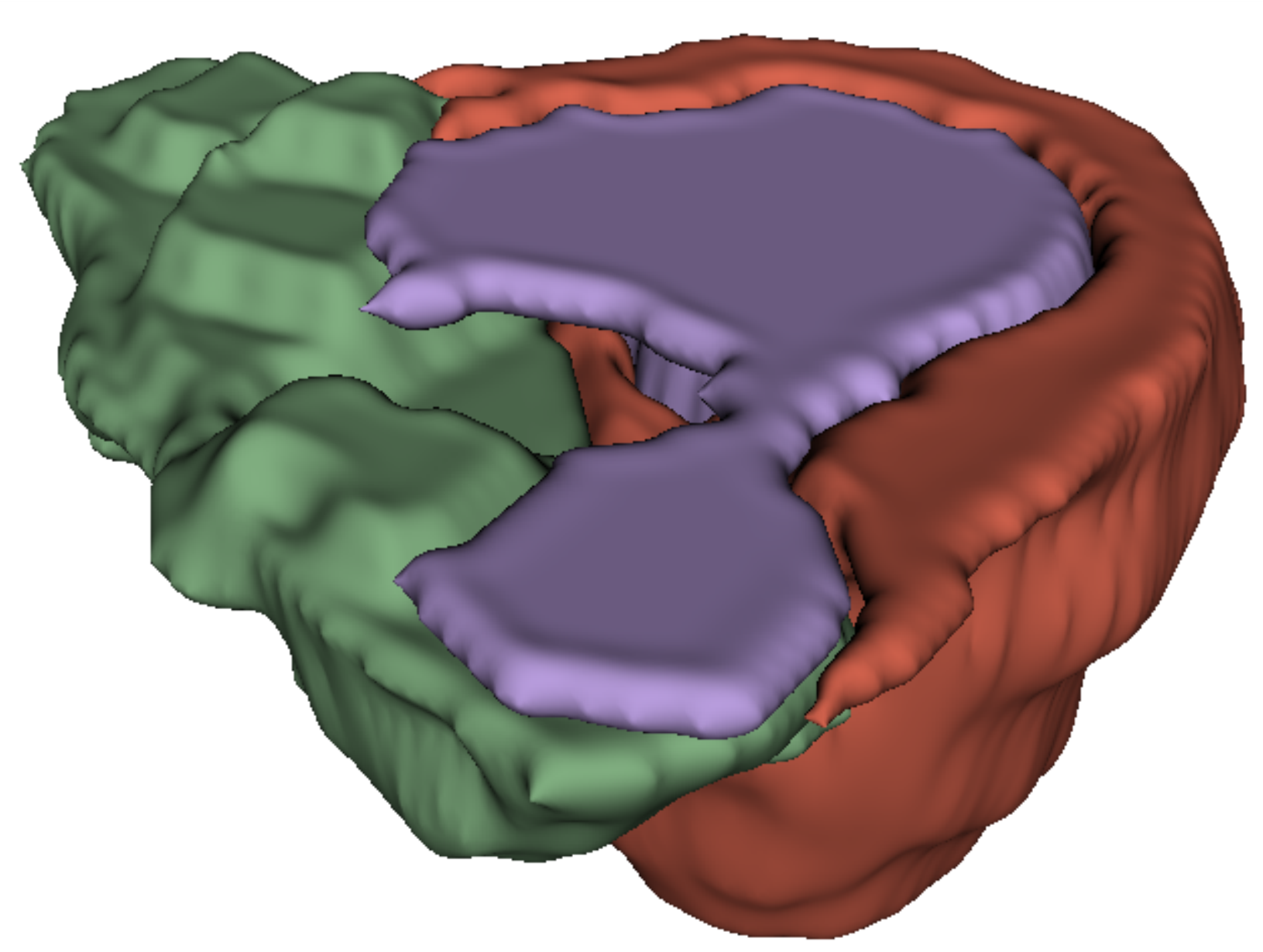}
        \caption{}
     \end{subfigure}
    \begin{subfigure}[b]{0.16\textwidth}
         \centering
         \includegraphics[width=\textwidth, height=0.08\textheight]{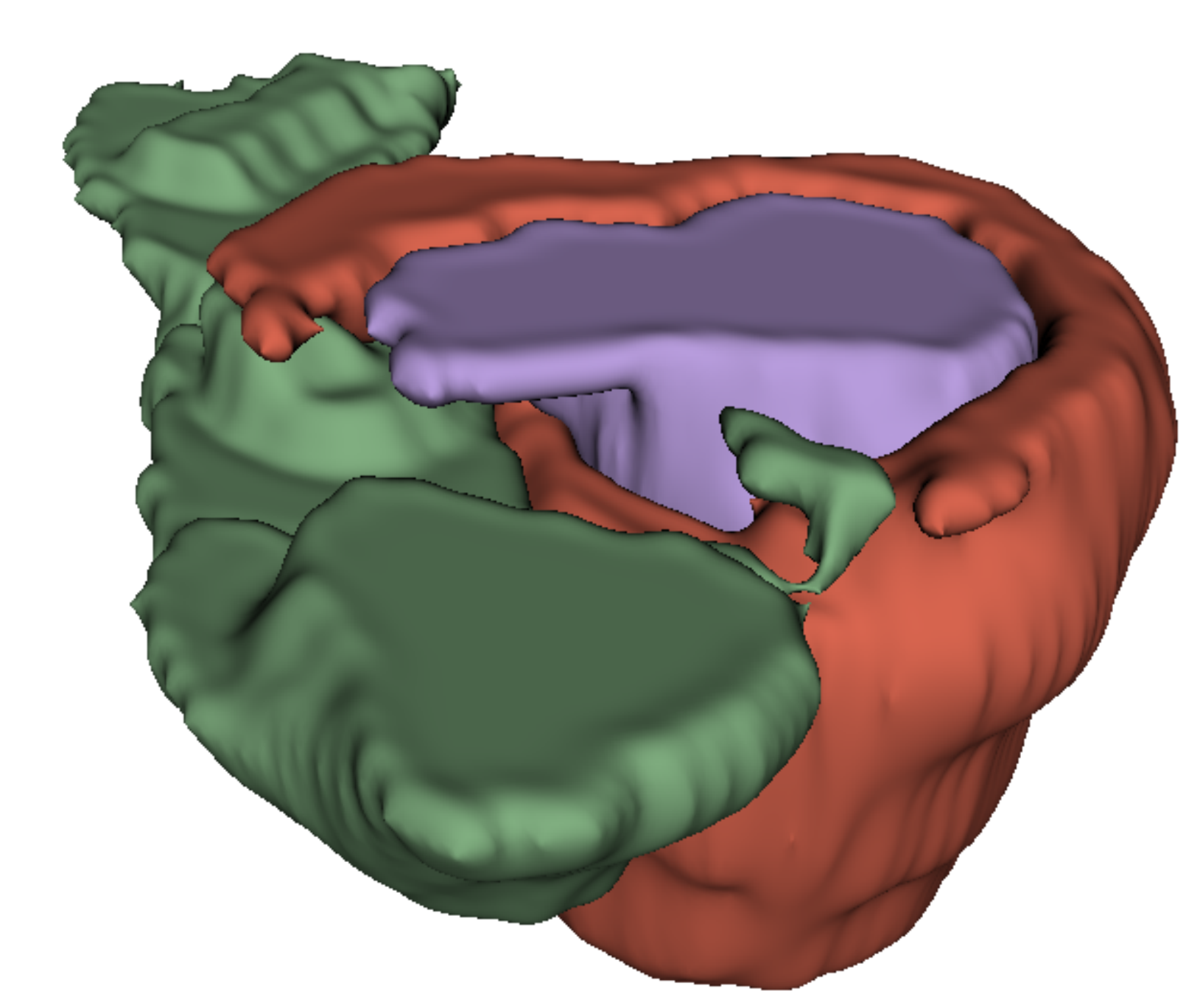}
        \caption{}
     \end{subfigure}
    \begin{subfigure}[b]{0.16\textwidth}
         \centering
         \includegraphics[width=\textwidth, height=0.08\textheight]{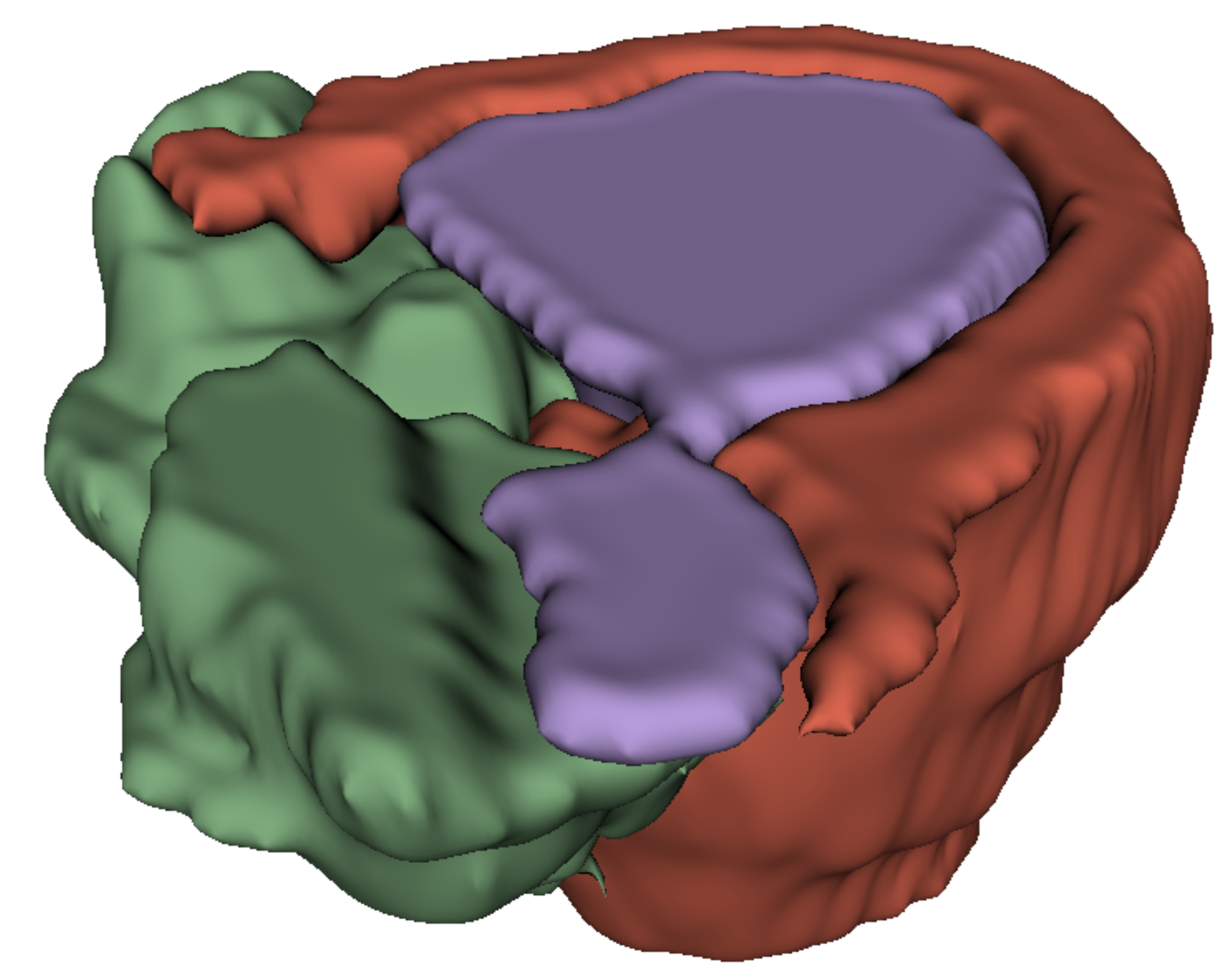}
        \caption{}
     \end{subfigure}
     \begin{subfigure}[b]{0.16\textwidth}
         \centering
         \includegraphics[width=\textwidth, height=0.08\textheight]{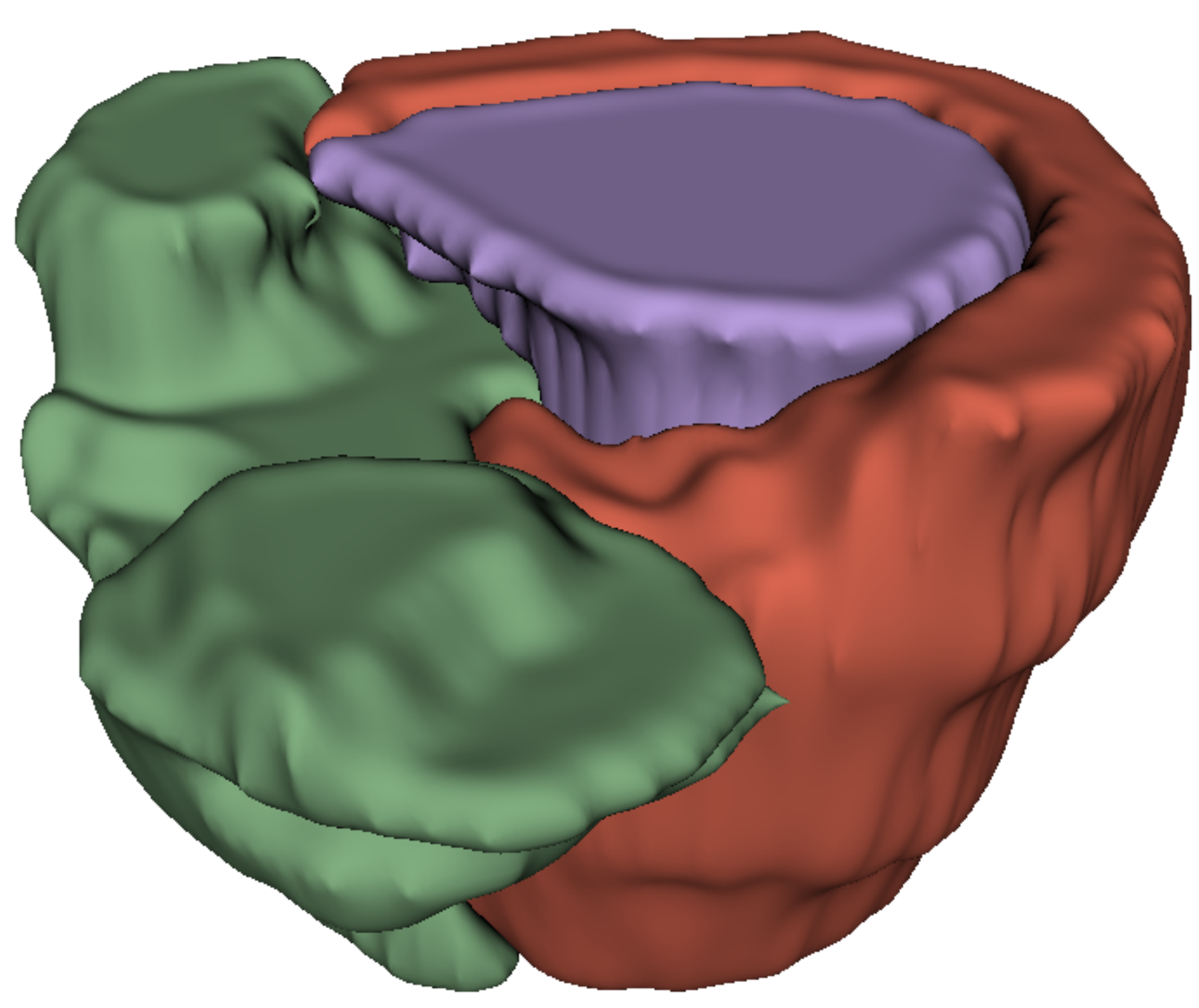}
        \caption{}
     \end{subfigure}
     
    \caption{3D Segmentation results under fewshot of ConFUDA for two LGE-CMRs from the test set using different components. (a) Ground-truth segmentation, (b) FUDA, (c) FUDA+CCL, (d) FUDA+CCL+CNR, (e) FUDA+CCL+CNR+MPCCL (ConFUDA).}
    \label{fig:ablationPlot}
\end{figure}

An ablation study is shown in \cref{tab:ablation} and \cref{fig:ablationPlot}. In all experiment settings, ConFUDA reaches the highest Dice with all the components and the lowest HD95 for oneshot and fewshot settings. In \cref{fig:ablationPlot}, from left to right, We observe a progressive improvement by adding components to the segmentation model. \cref{fig:ablationPlot} (\nth{2} row) shows that adding CCL corrects the misclassified RV to some extent. While apex slices are missing. By applying CNR, the apex slices are successfully predicted, and the predicted MYO structures are more complete. Further application of MPCCL removes the misclassified RV and MYO, resulting in the best prediction.

\begin{table}[!t]
\centering
\caption{Ablation study on all the components in the proposed ConFUDA in oneshot, fewshot, and full data experiment settings. Reported Dice and HD95 are average scores for all the organs.}
\label{tab:ablation}
\resizebox{\textwidth}{!}{%
\begin{tabular}{ccc|cc|cc}
\hline
Ablation study&\multicolumn{2}{|c}{\textbf{Oneshot}} & \multicolumn{2}{|c}{\textbf{Fewshot}} &\multicolumn{2}{|c}{\textbf{Fulldata}} \\ \hline
\multicolumn{1}{c|}{Methods}& \multicolumn{1}{c|}{Dice} & HD95& \multicolumn{1}{c|}{Dice} & HD95& \multicolumn{1}{c|}{Dice} & HD95\\ \hline
\multicolumn{1}{l|}{FUDA}& \multicolumn{1}{c|}{0.65$\pm$0.18}                         &12.5$\pm$19.4&\multicolumn{1}{c|}{0.67$\pm$0.17}&10.9$\pm$15.8&\multicolumn{1}{c|}{0.67$\pm$0.18}&8.7$\pm$9.9 \\
\multicolumn{1}{l|}{FUDA+CCL}& \multicolumn{1}{c|}{0.68$\pm$0.15} & 9.6$\pm$10.5& \multicolumn{1}{c|}{0.68$\pm$0.17}&11.0$\pm$14.8&\multicolumn{1}{c|}{0.69$\pm$0.11}&10.2$\pm$11.0  \\ 
\multicolumn{1}{l|}{FUDA+CCL+CNR} & \multicolumn{1}{c|}{0.68$\pm$0.16}& 9.5$\pm$11.3&\multicolumn{1}{c|}{0.71$\pm$0.14}&8.8$\pm$9.1&\multicolumn{1}{c|}{0.72$\pm$0.10}& \textbf{8.1$\pm$6.7}  \\ 
\multicolumn{1}{l|}{FUDA+CCL+CNR+MPCCL}      &\multicolumn{1}{c|}{\textbf{0.69$\pm$0.16}}&\textbf{9.0$\pm$10.4}&\multicolumn{1}{c|}{\textbf{0.72$\pm$0.13}}&\textbf{8.3$\pm$10.2}&\multicolumn{1}{c|}{\textbf{0.73$\pm$0.11}}&8.4$\pm$7.2  \\ \hline
\end{tabular}
}
\end{table}

\textbf{Discussion.} 
Despite the success of ConFUDA compared with other UDA methods,
we observe minor improvements after adding MPCCL to the model, especially for the full dataset. In some cases, we also observe an increase in HD95 by applying MPCCL. We attribute this problem to the batch size of the target, which is set to 1 in order to be compatible with the oneshot setting. For full data, during 200 epochs of training, each target sample can only be trained around five times, whereas for fewshot and oneshot it is 200$\sim$2000 times more. As a result, the model is undertrained in the target domain for the full dataset. In future work, we will try to solve this problem, and to further test the robustness of the proposed method, we aim to evaluate it on other multi-modal datasets.

\section{Conclusion}
In this work, we proposed a contrastive learning-based approach to tackle challenging fewshot UDA for medical image segmentation (\textbf{ConFUDA}). We equip ConFUDA with RAIN to generate diverse, stylized images and address the limited data problem. We further introduced an inter-and intra-contrastive loss computed on class-wise feature centroids of source and target data reducing the computational burden of contrastive learning. We also proposed CNR to control the magnitude of the target centroids, which further reduces the discrepancy between source and target centroids, and MPCCL to reduce the variance of the target features, leading to improved performance in cardiac image segmentation.

%
%
%
\newpage
\bibliographystyle{splncs04}
\bibliography{biblo.bib}
\end{document}